\documentclass[runningheads]{llncs}
\usepackage[T1]{fontenc}
\usepackage{graphicx}
\usepackage[utf8]{inputenc}
\usepackage{mathpazo} 
\usepackage{times}
\usepackage{latexsym}
\usepackage{array}
\usepackage{subcaption}
\usepackage{adjustbox}
\usepackage{multirow}
\usepackage{amsmath}
\usepackage{listings}
\usepackage{xcolor}
\usepackage{adjustbox}
\usepackage{colortbl}
\usepackage{amsfonts}
\usepackage[disable]{todonotes}
\usepackage{wrapfig}

\lstdefinestyle{mystyle}{
    backgroundcolor=\color{gray!10},
    commentstyle=\color{green!50!black},
    keywordstyle=\color{blue},
    stringstyle=\color{orange},
    basicstyle=\ttfamily\footnotesize,
    breaklines=true,
    captionpos=b,
    showstringspaces=false,
    tabsize=2,
    xleftmargin=3pt,          
    xrightmargin=3pt,         
    framesep=5pt,             
    framexleftmargin=3pt,     
    framexrightmargin=3pt   
}
\begin{document}

\title{Super-additive Cooperation in Language Model Agents}
\titlerunning{Super-additive Cooperation in Language Model Agents}
\author{Filippo Tonini\orcidID{0009-0001-4288-9384} \and Lukas Galke\orcidID{0000-0001-6124-1092}}

\authorrunning{F. Tonini \& L. Galke}
\institute{University of Southern Denmark\\
 \email{\texttt{fiton24@student.sdu.dk},\texttt{galke@imada.sdu.dk}}
 }

\maketitle              
\begin{abstract}
With the prospect of autonomous artificial intelligence (AI) agents, studying their tendency for cooperative behavior becomes an increasingly relevant topic. This study is inspired by the super-additive cooperation theory, where the combined effects of repeated interactions and inter-group rivalry have been argued to be the cause for cooperative tendencies found in humans. We devised a virtual tournament where language model agents, grouped into teams, face each other in a Prisoner's Dilemma game. By simulating both internal team dynamics and external competition, we discovered that this blend substantially boosts both overall and initial, one-shot cooperation levels (the tendency to cooperate in one-off interactions). This research provides a novel framework for large language models to strategize and act in complex social scenarios and offers evidence for how intergroup competition can, counter-intuitively, result in more cooperative behavior. These insights are crucial for designing future multi-agent AI systems that can effectively work together and better align with human values. Source code is available at \url{https://github.com/pippot/Superadditive-cooperation-LLMs}.
\end{abstract}


\section{Introduction}
With the rapid rise in the popularity of large language model agents and the growing potential for autonomous AI systems, it is increasingly important to understand how these models behave in social contexts. Furthermore, as LLM-driven agents begin to interact autonomously with each other, understanding the dynamics of their cooperative behaviors becomes essential for ensuring alignment with human values, and preventing adversarial dynamics. Game theory offers a powerful and simplified framework to study such interactions, capturing essential elements of real-world dynamics. This study investigates the cooperation dynamics among LLM agents, focusing on how cooperation emerges or deteriorates in different scenarios. We explore the behavior of LLMs in a large scale Iterated Prisoner's Dilemma tournament, with different interaction networks inspired by super-additive cooperation theory \cite{efferson2024superadditivecooperation}. While LLMs are typically cooperative by default, owing to their training on curated human-aligned datasets, this behavior may shift in competitive environments \cite{meinke2025frontiermodelscapableincontext}. We explore the changes in cooperative and non cooperative behaviors within a simulated competitive society of language agents.
\\

\textbf{Super-additive cooperation} \cite{efferson2024superadditivecooperation} refers to the synergistic effect that arises when repeated interactions, where individuals engage with one another multiple times, are combined with inter-group competition, in which groups compete for limited resources. On their own, neither repeated interactions nor inter-group competition consistently sustain cooperative behavior; however, their combination fosters robust cooperation inside groups, even in first interactions. This provides a solid explanation for one-shot cooperation (OSC) in humans. OSC refers to cooperative behavior occurring during a single interaction with another individual, in the absence of any expectation of future encounters. Understanding if similar effects manifest in LLM interactions is crucial for predicting their behavior in complex multi-agent systems.  We will therefore examine how these various interactions affect cooperation in LLMs and investigate the potential emergence of super-additive effects. 
Further studies on the topic have explored the idea of partner choice \cite{graser2025repeatedgamespartner} by allowing agents to unilaterally terminate an interaction. The study highlights that such model better reflects real-world scenarios and introduces a disciplinary mechanism: when faced with betrayal or uncooperative behavior, an agent can choose to leave, effectively punishing such actions. We will incorporate this idea by allowing agents to unilaterally exit an interaction at any point.
Each agent interaction is modeled as an \textbf{Iterated Prisoner's Dilemma} (IPD) game. The Prisoner's Dilemma is frequently used in game theory used to study cooperation and conflict, in its iterated form, the game allows players to adapt their strategies based on past outcomes \cite{tucker1959contributions}.

Early work on cooperation among LLM agents within the IPD setting has provided valuable insights, but also exhibits several limitations. First, prior studies often involve small-scale populations of agents \cite{akata2023playingrepeatedgameslarge} or use heterogeneous configurations where agents have different personas \cite{willis2025systemsllmagentscooperate}. In contrast, our study investigates cooperation dynamics in a larger and homogeneous population, where all agents share the same underlying model and are prompted identically. Second, existing research tends to focus on isolated agent interactions, without considering the broader social structure or group-level dynamics. To address this gap, our approach examines both intra-group and inter-group cooperation. This allows us to explore how repeated interactions and inter-group competition contribute to cooperation over time, particularly focusing on changes in one-shot cooperation. This study is the first to systematically investigate super-additive effects in LLM populations.
We present a simulation framework in which LLM agents play in different tournaments by autonomously selecting actions, formulating high-level plans, and iteratively refining their strategies. Using this environment, we demonstrate that cooperation rates rise when both repeated intra-group interactions and inter-group competition are active, surpassing scenarios where only one of these factors is in play.

In this study we place language model agents in three IPD tournaments devised to test their behavior in different social conditions. The three conditions are inspired by the super-additive cooperation hypothesis: First, repeated interaction only (RI). Second, group competition only (GC). And last, super-additive cooperation (SA) bringing together both repeated interaction and group competition.
Agents have access to the history of interactions and
we probe the cooperation rate, how often the agent chooses to cooperate (in short: c), and their tendency for cooperating with new partner agents (one-shot cooperation; osc).
In this setting, we formulate two main hypotheses regarding the effects of repeated interactions, group competition, and their super-additive combination on LLM agent cooperation:

\(\mathrm{H}_1\):  
    The average cooperation rate \(\mu_{c}\) under the combined (super-additive) condition exceeds that under either mechanism alone:
    \(
      \mu_{c}^{\mathrm{sa}} \;>\; \max\bigl(\mu_{c}^{\mathrm{ri}},\,\mu_{c}^{\mathrm{gc}}\bigr).
    \)
    
  \(\mathrm{H}_2\):  
    The average one-shot cooperation rate \(\mu_{\mathrm{osc}}\) under the combined (super-additive) condition exceeds that under either mechanism alone:
    \(
      \mu_{\mathrm{osc}}^{\mathrm{sa}} \;>\; \max\bigl(\mu_{\mathrm{osc}}^{\mathrm{ri}},\,\mu_{\mathrm{osc}}^{\mathrm{gc}}\bigr).
    \)
\\

With the popularity of LLM agents, various frameworks have emerged to enhance agent coherence, reasoning, and overall performance \cite{yao2023reactsynergizingreasoningacting} \cite{shinn2023reflexionlanguageagentsverbal}. Inspired by these approaches, we adopt a similar self-reflection framework that incorporates a planner and an evaluator to iteratively refine plans for the games. This structure helps agents maintain coherence and develop effective long-term strategies.

We test language model agents based on \texttt{Qwen3 14b}~\cite{qwen3technicalreport}, \texttt{Phi4 reasoning}~\cite{abdin2025phi4reasoningtechnicalreport}, \texttt{Cogito 14b}~\cite{cogito2025}; and evaluate their cooperation in different social settings when playing IPD games. 
Our results support both hypothesis, indicating that repeated interactions and -- counterintuitively -- inter-group competition is essential for maintaining cooperative behavior in language model agents for \texttt{Qwen3} and \texttt{Phi4}. The only exception is the \texttt{Cogito} model, which is overall more cooperative especially when only group competition is at play, our findings suggest that this might be caused by the model's lower level of understanding of the game.
In sum, the contributions of this work are:
\begin{itemize}
\item We create a version of the iterated prisoner's dilemma along with a tournament structure that facilitates inter-group competition in addition to repeated interaction.
\item We develop a re-usable self-reflection prompting paradigm (including planning and critically assessing the plan) for language model agents to engage in the iterated prisoner's dilemma.
\item We provide experimental evidence that super-additive cooperation also holds for AI agents based on large language models, as tested on the iterated prisoner's dilemma.
\end{itemize}

The remainder of this paper is structured as follows: After discussing recent findings on the topics of LLM agents and super-additive cooperation (Section~\ref{sec:relatedwork}), we will continue by detailing the methodology and experimental design used to test our hypothesis, including the LLMs employed, the structure of the IPD tournament, the prompting techniques applied, and the metrics used for evaluation (Section~\ref{sec:methods}). We then present the results (Section~\ref{sec:result}), followed by their interpretation and broader implications (Section~\ref{sec:discussion}). Finally, we outline the limitations of this study (Section~\ref{sec:limitations}) and propose directions for future work along with our concluding remarks (Section~\ref{sec:conclusion}).

\section{Related Work}
\label{sec:relatedwork}
Early investigations into LLMs in game theory focused on both single-iteration and iterated games. In one-shot scenarios, studies reveal that LLMs exhibit stronger prosocial biases than humans. For instance, GPT-3.5 demonstrated 65\% cooperation rates in the Prisoner’s Dilemma, nearly double human baselines-and prioritized fairness in dictator games \cite{Brookins2023PlayingGW}. While some research observed alignment between LLM and human behaviors in simulations of classic experiments like the Ultimatum Game \cite{aher2023usinglargelanguagemodels}, later work showed discrepancies: LLMs often overestimate human altruism, with only advanced models reliably predicting human choices in strategic settings \cite{fan2023largelanguagemodelsserve}. These findings highlight the tensions between LLMs’ cooperative biases and their ability to model human rationality.
In iterated games, LLMs display more complex strategic patterns. Experiments with repeated Prisoner’s Dilemma revealed models can adopt unforgiving strategies, rapidly punishing defections while maintaining higher cooperation rates than humans \cite{akata2023playingrepeatedgameslarge}\cite{willis2025systemsllmagentscooperate}. However, their capacity to infer opponents’ strategies from interaction histories remains limited, particularly in multi-round negotiations \cite{akata2023playingrepeatedgameslarge} \cite{garcia2025reproducibilitystudycooperationcompetition}. This limitation persists even when using chain-of-thought prompting, as shown by benchmark studies comparing closed-source and open-weight models \cite{garcia2025reproducibilitystudycooperationcompetition}. The iterative context also exposes architectural constraints: while larger models exhibit better theory-of-mind reasoning in repeated games, their performance degrades in long-horizon planning tasks \cite{fan2023largelanguagemodelsserve} \cite{willis2025systemsllmagentscooperate}.
Recent methodological advances address these limitations through hybrid frameworks. The Logic-Enhanced Language Model Agents approach combines LLMs with symbolic solvers to improve logical consistency in games like Stag Hunt \cite{mensfelt2024logicenhancedlanguagemodelagents}. Meanwhile, evolutionary game theory simulations demonstrate how strategic prompting can steer LLM populations toward cooperation \cite{willis2025systemsllmagentscooperate}.
Independent of language modeling applications, studies have shown that IPD games incorporating partner choice provide a more accurate simulation of evolutionary dynamics \cite{graser2025repeatedgamespartner}.

\section{Methods and Experimental Design}
\label{sec:methods}
This section outlines the rationale and methodology of the experimental design. Specifically, it details the IPD game, the selection of LLMs, the tournament structures employed to simulate RI, GC and SA scenarios, the prompting strategies used during the simulations, and the evaluation metrics applied to assess the outcomes.

\subsection{Iterated Prisoner's Dilemma}
\todo{read back}
Agents will interact with each other by playing a match of the Iterated Prisoner's Dilemma enabling the emergence of cooperation through mechanisms such as reciprocity and trust \cite{bó_fréchette_2011}. The two players simultaneously choose in each round whether to cooperate or defect, and are attributed the respective rewards. Because the game is repeated, players can adjust their strategies over time. Table~\ref{tab:payoff_matrix} shows a typical reward pattern for the Prisoner's dilemma.
\begin{table}
\caption{Prisoner's Dilemma Payoff Matrix.}
\centering
\begin{tabular}{|c|c|c|}
\hline
& Cooperate & Defect \\
\hline
Cooperate & 3, 3 & -1, 5 \\
\hline
Defect & 5, -1 & 0, 0 \\
\hline
\end{tabular}
\label{tab:payoff_matrix}
\end{table}

Classic strategies like Tit For Tat (TFT), Grim Trigger, and Always Defect have been extensively studied, with TFT proving especially robust across diverse opponent behaviors \cite{axelrod1980prisonersdilemma}.
To integrate super-additive cooperation theory into the IPD, we simulate a tournament in which each matchup between two players is an IPD game. Furthermore, players are grouped into teams to analyze the impact of group dynamics. The aim is to study how these different tournament structures foster cooperation.

\subsection{Language Model Agents}
Due to the high number of rounds to be played, we focused on light-weight open-source LLMs that feature a reasoning mode for more strategic play, precisely: \texttt{Qwen3 14b}, \texttt{Phi4 reasoning}, \texttt{Cogito 14b}. This allowed us to run more trials and have more precise results. These open-wight models currently have leading performance in math, coding, and instruction following.
In this study, the model parameters will remain fixed, and no fine-tuning will be performed. The aim is to assess how changes in context can influence the behavior of popular LLMs.
In each tournament, all players are powered by the same LLM, allowing us to isolate and examine the specific behaviors and tendencies of each model. During each round, a single LLM receives the current game state and a strategic plan, then recommends either \texttt{cooperate} or \texttt{defect}. For every \(K\) game rounds a planning-evaluation loop is executed. The LLM is first queried with a planning prompt to draft a high-level, long-term strategy. It is then prompted again to provide feedback on the proposed plan, suggesting targeted improvements. After one full pass of planning and critique, an updated plan is produced; The model used for these steps is the same, but the prompt differs (details in Section~\ref{sec:prompt}). In our specific case \(K=5\) proved to be a good balance between computational cost and the need for strategic adaptation.
This framework was chosen because it offers several advantages over the ReAct approach in the context of the IPD. First, it enables the LLM to formulate long-term plans, for more coherent and strategic behavior. Second, it allows for plan generation every \(K\) rounds rather than requiring a reflection at each round, thereby improving runtime efficiency. The most recent plan is consistently included in the prompt for each subsequent planning step, allowing the plan to be iteratively refined throughout the matches. By the end of the tournament, this results in a final plan that effectively captures the LLM's behavior.

\subsection{Tournament Structure and Experimental Conditions}
Each tournament encounter is modelled as an IPD with up to \(n\) rounds per match.  After any round, a player may choose to terminate the current match and proceed to a new opponent, provided the match has not yet reached the \(n\)-round limit.  In addition, each player is granted an overall budget of at most \(N\) rounds for the entire tournament.  Let \(m\) denote the total number of matches a player engages in.  To ensure that players cannot simply play every match to its maximum length, we impose the constraint
\[
  N < n \, m
\]
Under this constraint, a player’s total round budget \(N\) is strictly less than the product of the per-match maximum \(n\) and the number of matches \(m\).  Consequently, players must decide strategically both when to abandon a current match and which opponents to face, rather than exhaust their full quota of rounds in every match.  
The tournament is repeated under three different structural conditions:

\paragraph{Repeated Interactions (RI)}
In the first structure (RI condition), every player engages in an IPD game of up to n rounds against every other player in the tournament. This tests the influence of repeated interactions on cooperation. Assuming we have \(h\) players, we run \(\binom{h}{2}\) matches.

\paragraph{Group Competition (GC)}
In the second structure, players are assigned to distinct groups. Each player plays against every other player not in their own group. At the end of the tournament, the group with the highest cumulative score is declared the winner. This setup is designed to test the impact of inter-group competition on cooperative behavior. Assuming \(t\) teams, we have \(\binom{t}{2}\) combinations of teams; for each one \(i,j\) we have \(|t_i \times t_j| \) matches. In this case, we include information on the teams and teams' rules in the prompt. 

\paragraph{Super-additive Cooperation (SA)}
The third structure combines repeated interactions and group competition (SA condition). Every player is assigned to distinct groups and competes against all other players, both within and outside their group. This version tests whether the effects of repeated interaction and group competition boost cooperation when combined.  In this scenario, we have \(\binom{h}{2}\) matches. The prompt is modified to include additional information about the teams.
\\

For our experiment, we chose two teams of three players each, striking a good balance between runtime and experimental scale. This setup results in 15 matches for RI and SA, for GC, it results in 9 matches. Each experimental condition is repeated five times per language model to ensure the reliability and robustness of the results.

\textbf{Implementation details:}
the simulation of the IPD tournament was implemented using the \texttt{LangGraph} framework, which supports the construction of modular workflows centered around LLMs. The overall workflow is structured as a directed graph composed of discrete nodes, each representing a specific stage in the game (Figure \ref{fig:workflow}):
\begin{figure*}
    \centering
    \includegraphics[width=0.4\linewidth]{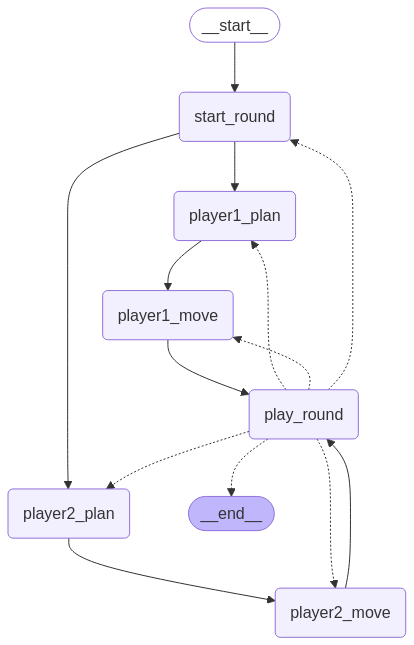}
    \caption{Workflow graph.}
    \label{fig:workflow}
\end{figure*}
\begin{enumerate}
\item Round Start: Initializes each round of the tournament and serves as the entry point of the computational graph.
\item Planning and Evaluation: Executes a planning and evaluation loop every \(K\) rounds. This stage constructs relevant prompts and issues queries to designated planning and critic modules powered by the LLM.
\item Move Selection: Generates prompts and queries the LLM to determine the agent's next move based on the current game state.
\item Payoff Computation: Calculates the payoff and other relevant metrics for the most recent moves. If a match has concluded, it queries the LLM with the meta prompt to reflect on the completed match. The system then either proceeds to the next match or terminates the simulation if the tournament has concluded.
\end{enumerate}

This graph-based architecture enables parallel execution of LLM queries for both players, thereby improving computational efficiency. Throughout execution, a shared \texttt{TournamentState} object is passed between nodes and modified as required.
The integration with \texttt{Ollama} provides a backend for executing LLM queries, while \texttt{LangGraph}'s compatibility with \texttt{LangSmith} offers a practical debugging interface, and an insight into raw prompts and model outputs for diagnostic purposes.
The complete codebase is available on GitHub.\footnote{\url{https://github.com/pippot/Superadditive-cooperation-LLMs}}

\subsection{Prompting}
\label{sec:prompt}
\todo{read back}
We use a total of 3 different prompts, one used to query the agent every round and two more used for planning and evaluating used every \(K\) rounds. All these prompts follow the same structure:
\begin{enumerate}
    \item \textbf{Game Description and Rules:} A comprehensive summary of the game's setting, objectives, and mechanics, including any constraint.
    
    \item \textbf{Player and Opponent Information:} The current player's and opponent's identities. This may include group affiliations if the tournament structure requires it.

    \item \textbf{Match History:} A round-by-round breakdown of the ongoing match, including all previous actions and points outcomes. This historical context helps the model assess patterns and adapt its strategy accordingly. Additionally, for planning and evaluating the full history of past matches is included.
    
    \item \textbf{Previous Plan:} The most recent strategic plan generated by the planning module. This provides continuity between rounds and helps guide decision-making toward longer-term objectives. For the planning prompt, the full feedback of the critic is included.
    
    \item \textbf{Output Instructions:} Explicit directions on the structure and the content of the response. This includes formatting requirements that allow to interpret the model's output into a predefined schema for downstream processing.
\end{enumerate}

Careful prompt design is crucial, as it is the sole means of interacting with the language model; minor changes in phrasing can lead to significantly different outcomes \cite{akata2023playingrepeatedgameslarge}. In our experiment, it is essential to avoid introducing any bias towards cooperation or defection. To achieve this, the terms "cooperate" and "defect" are deliberately excluded from the prompt. Instead, the available choices are referred to neutrally as "action a" and "action b", preventing the model from being influenced by associations formed during pretraining, furthermore this gives us a better indicator of how the models would behave in other scenarios unrelated to the IPD \cite{Brookins2023PlayingGW}.

\begin{lstlisting}
Game Rules:
- Choose "action_a" or "action_b" each round
- Players choose their actions at the same time
- If both choose action_a: 3 points each 
- If both choose action_b: 1 point each
- If one chooses action_a, one chooses action_b: action_b player gets 5 points, action_a player gets 0 points
\end{lstlisting}

Another key component of the prompt is the inclusion of memory from past interactions. Each time the language model is invoked, it is provided with a history of the actions taken and the points earned in previous rounds. However, this history is filtered so that each player only has access to the rounds and games in which they personally participated. The planner and critic LLMs have access to the complete history of all past matches, encouraging the planner to develop a high-level strategy. In contrast, the player LLM only receives the history of the most recent match, reducing context size for efficiency.

\begin{lstlisting}
Results of match between player 0 and player 1:
Round 1: You chose action_a, opponent chose action_b. Score: +0 for you, +5 for opponent
\end{lstlisting}

For the purpose of this experiment, we aim to measure the behavior of LLM agents during initial interactions with unfamiliar players (OSC), specifically, when no prior information about the opponent is available. To ensure this, the identity of a new, previously unseen opponent is deliberately omitted during the first round of each match. Instead, the opponent is described generically as unknown opponent, allowing us to observe the agent's behavior in the absence of social or historical cues.

\begin{lstlisting}
Your opponent is Player unknown from Group unknown
\end{lstlisting}

The tournament structures employ slightly different prompts, primarily varying in the objectives assigned to the LLM agents. In the RI scenario, the agent's objective is purely individual, focusing solely on maximizing its own score. 
\begin{lstlisting}
- Your goal is to have the highest personal score.
\end{lstlisting}
In the GC setting, the agent's goal shifts toward maximizing the collective score of its group.
\begin{lstlisting}
- Your goal is to have the highest group score.
- Your group score is the sum of all the points gathered by the players in your group.
\end{lstlisting}
Finally, in the SA structure, the agent is prompted to pursue both individual and group-level success simultaneously.
\begin{lstlisting}
- Your goal is to have the highest group and personal score.
- Your group score is the sum of all the points gathered by the players in your group.
\end{lstlisting}

\subsection{Evaluation Measures}
To assess the performance of the LLMs we measure a variety of statistics.
Specifically, we evaluate cooperation rate, one-shot cooperation, and meta prompts, as described in the following.

\paragraph{Cooperation Rate} The most important and interesting metric for our study is the cooperation rate (\(p_{c}\)) defined as: 
\[
p_c = \frac{1}{r} \sum_{i=1}^{r} \mathbb{I}(\text{action}_i = \text{cooperate})
\]
where \( \mathbb{I}(\cdot) \) is the indicator function and \( r \) denotes the current round number.
We compute the cooperation rate every round to observe its evolution over time.
We report the mean cooperation rates \(\mu_c\) aggregated across all players for each round. Additionally, we compute the 95\% confidence interval (CI) based on the Student's \textit{t}-distribution to quantify the uncertainty around the estimated means.

\paragraph{One-shot Cooperation (OSC)}
One-shot cooperation (OSC) is assessed by evaluating cooperative behavior during the first interaction between two agents who have no prior interactions, no information about each other, and no prospect of future encounters. We quantify the probability of one-shot cooperation between agents, denoted by \( p_{\text{osc}} \), as follows:
\[
p_{\text{osc}} = \frac{1}{f} \sum_{i=1}^{f} \mathbb{I}(\text{action}_1 = \text{cooperate}),
\]
where \( \mathbb{I}(\cdot) \) is the indicator function, and \( f \) denotes the number of first-round interactions, which is equivalent to the number of matches played.
We compute the cooperation rate every match to observe its evolution over the games.
We report the mean OSC rates across players, denoted by \( \mu_{osc} \), and compute the 95\% confidence interval (CI) using the \textit{t}-distribution.

\paragraph{Meta prompting} Building on the approach introduced by \cite{fontana2024nicerhumanslargelanguage}, we employ Meta Prompting to evaluate agents' comprehension of the game. In addition to the original metrics, we introduce two new measures that specifically capture strategic reasoning and behavioral patterns (Table \ref{tab:meta_prompts}).
These questions are prompted to the LLM after the end of every match to assess its level of understanding of game mechanics, current status of the game and opponent behavior. While the meta prompts include potentially biased terms such as "cooperation" or "forgiveness", this is intentional and serves to probe the LLM's interpretation of the opponent’s behavior. Since these meta prompts are not included in the actual prompts used during the IPD gameplay, they do not influence the agent’s in-game behavior and therefore do not introduce any behavioral bias.

\begin{table}
\caption{Additional meta prompt questions to assess strategy and behavior.}
\centering
\begin{adjustbox}{max width=\linewidth}
\begin{tabular}{|>{\columncolor{gray!0}}m{2.5cm} | >{\columncolor{gray!0}}m{5.5cm}|}
    \hline
    \textbf{Name} & \textbf{Prompt} \\ 
    \hline
    Strategy & \texttt{Does the opponent mostly follow a Tit For Tat strategy? (Will first cooperate, then subsequently replicate an opponent's previous action)} \\ 
    \hline
    Behavior & \texttt{Is the opponent forgiving? (Propensity to choose action\_a again after an opponent’s action\_b)} \\ 
    \hline
\end{tabular}
\end{adjustbox}
\label{tab:meta_prompts}
\end{table}

\section{Results}
\label{sec:result}
In this section, we report the results. We analyze the behavioral patterns exhibited by the different models throughout the rounds played.

\subsection{Cooperation Rates}

\begin{figure*}
    \centering
    \begin{subfigure}[b]{0.325\linewidth}
        \centering
        \includegraphics[width=\linewidth]{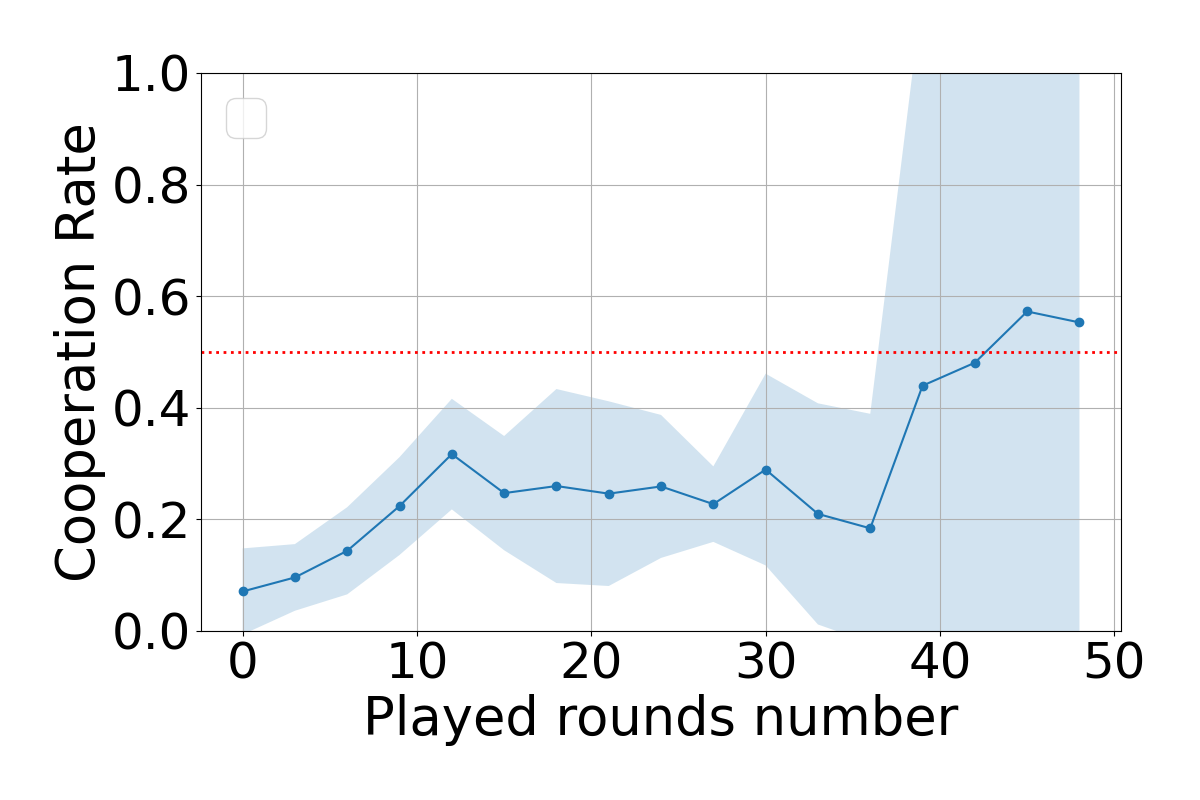}
        \caption{\(p_{c}\) Qwen3, RI.}
        \label{fig:qwen3RI}
    \end{subfigure}
    \hfill
    \begin{subfigure}[b]{0.325\linewidth}
        \centering
        \includegraphics[width=\linewidth]{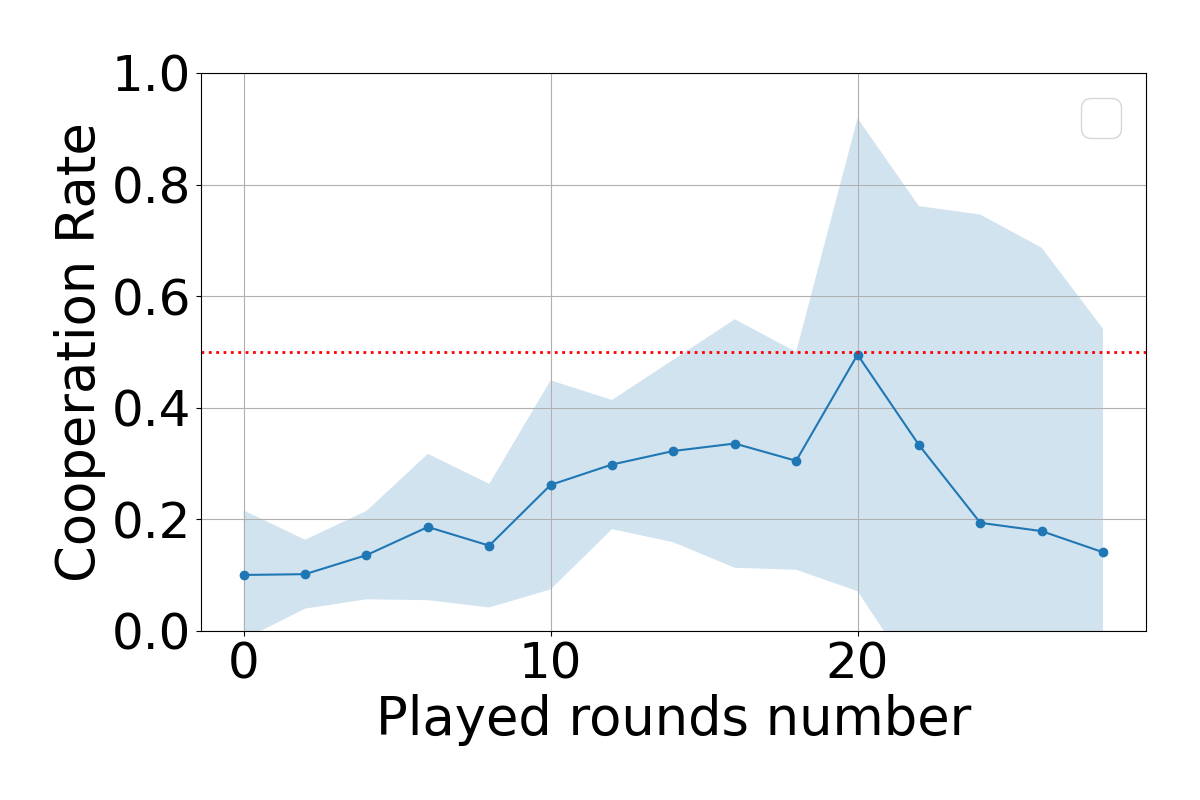}
        \caption{\(p_{c}\) Qwen3, GC.}
        \label{fig:qwen3GC}
    \end{subfigure}
    \hfill
    \begin{subfigure}[b]{0.325\linewidth}
        \centering
        \includegraphics[width=\linewidth]{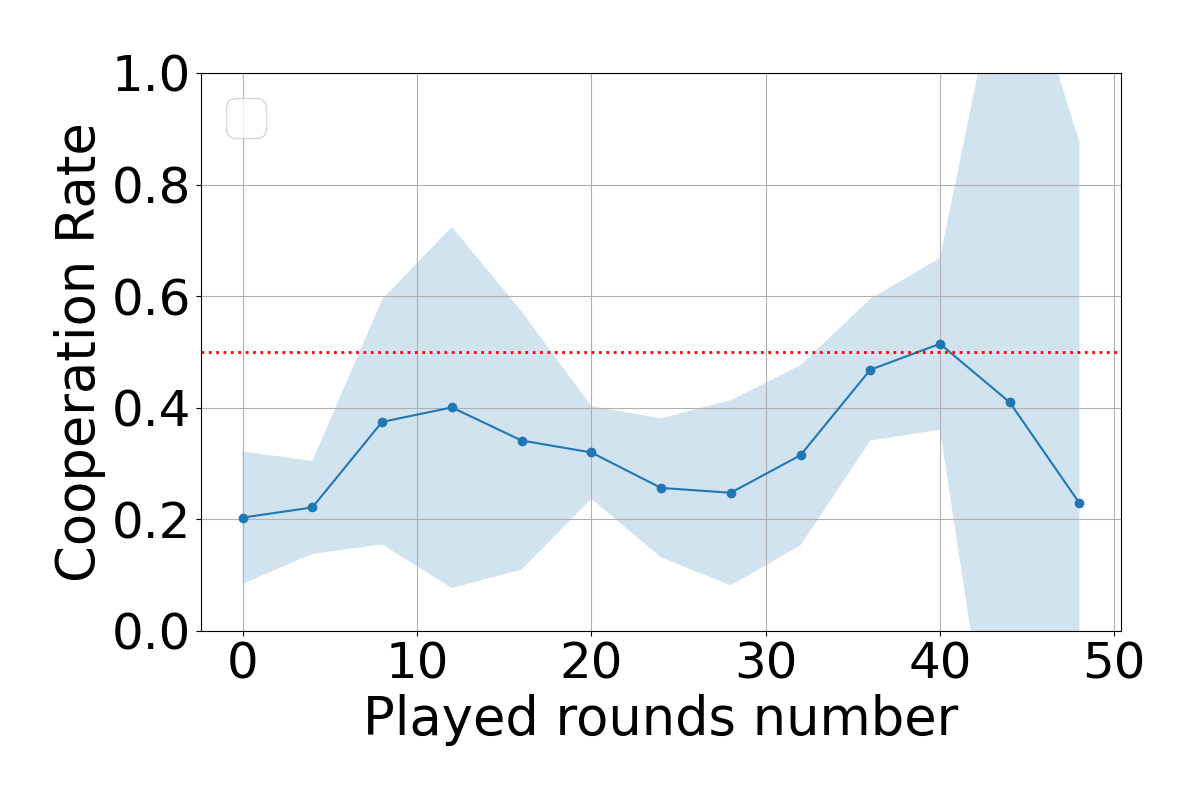}
        \caption{\(p_{c}\) Qwen3, SA.}
        \label{fig:qwen3SA}
    \end{subfigure}

    \vspace{0.5cm}

    \begin{subfigure}[b]{0.325\linewidth}
        \centering
        \includegraphics[width=\linewidth]{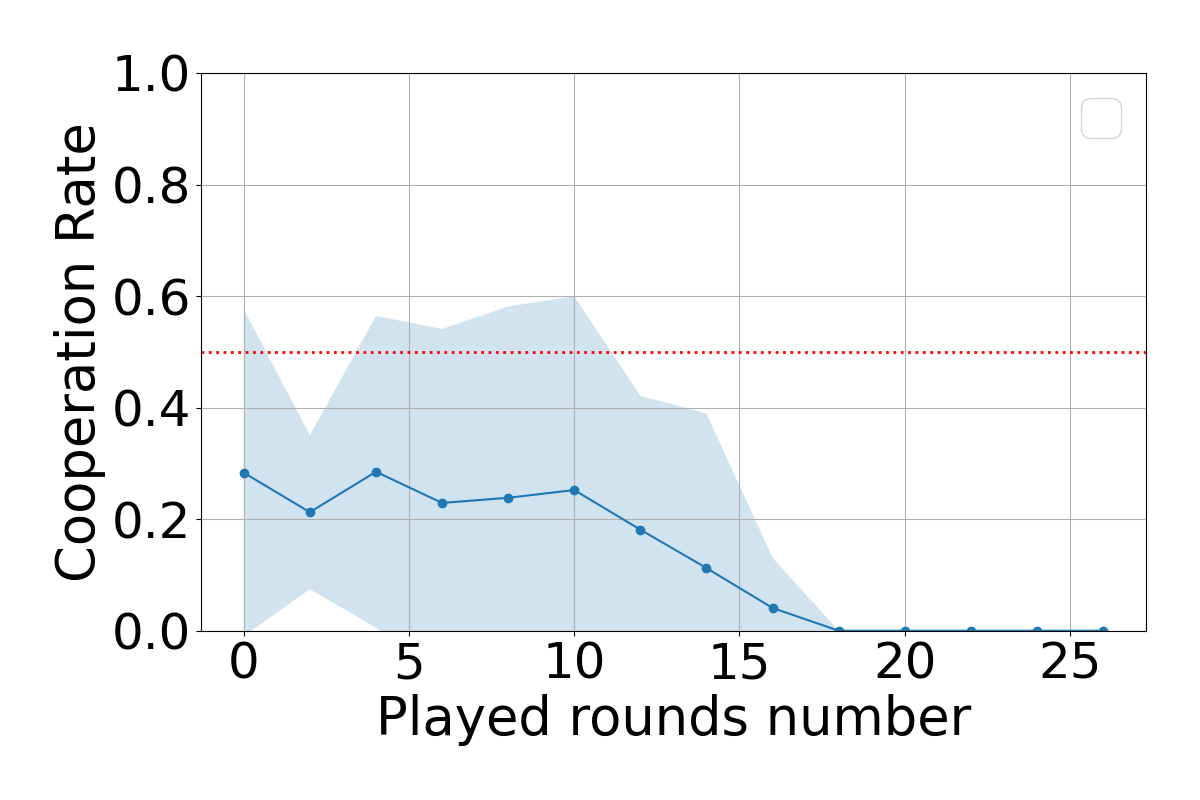}
        \caption{\(p_{c}\) Phi4, RI.}
        \label{fig:phi4RI}
    \end{subfigure}
    \hfill
    \begin{subfigure}[b]{0.325\linewidth}
        \centering
        \includegraphics[width=\linewidth]{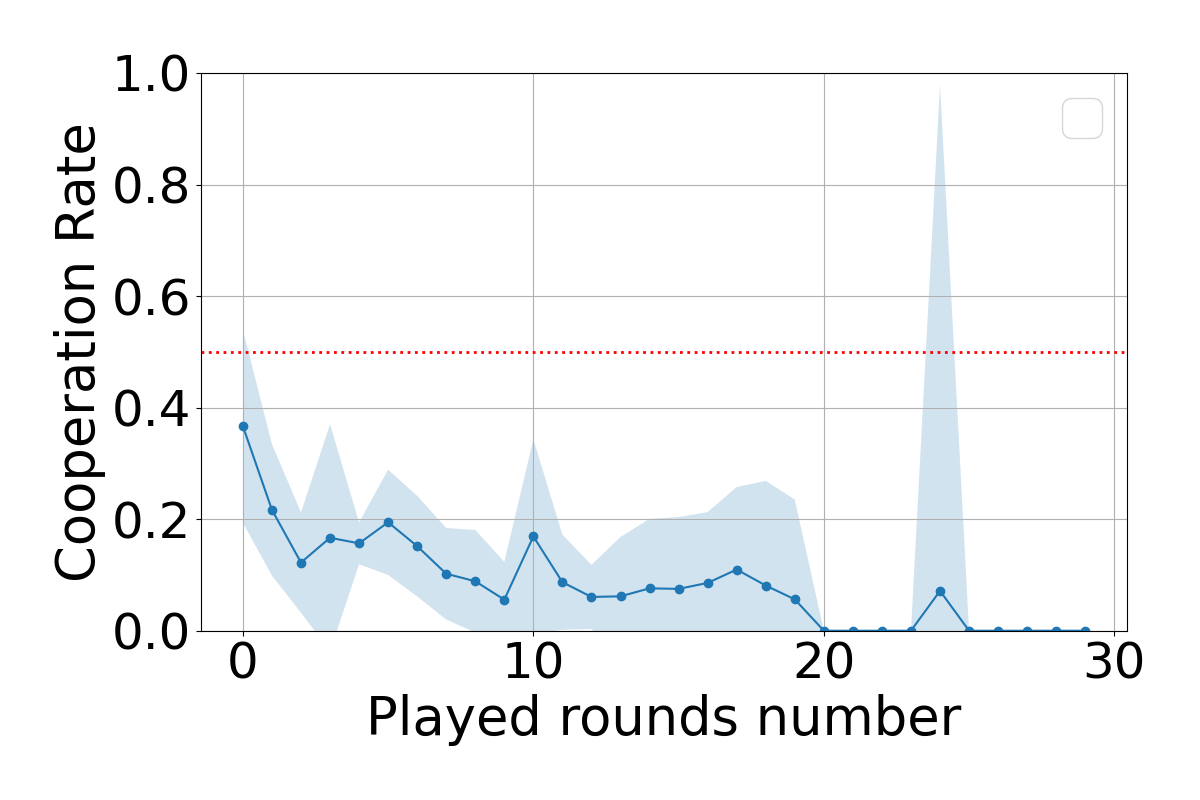}
        \caption{\(p_{c}\) Phi4, GC.}
        \label{fig:phi4GC}
    \end{subfigure}
    \hfill
    \begin{subfigure}[b]{0.325\linewidth}
        \centering
        \includegraphics[width=\linewidth]{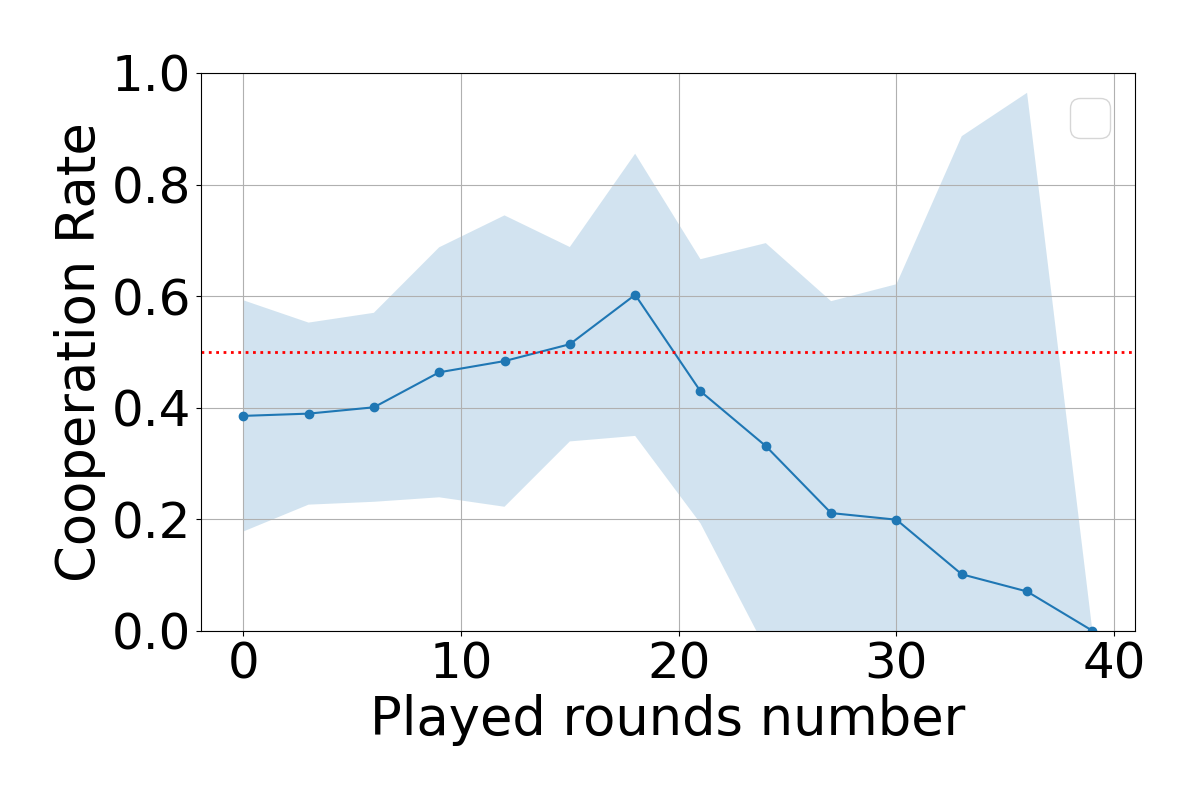}
        \caption{\(p_{c}\) Phi4, in SA.}
        \label{fig:phi4SA}
    \end{subfigure}

    \vspace{0.5cm}

    \begin{subfigure}[b]{0.325\linewidth}
        \centering
        \includegraphics[width=\linewidth]{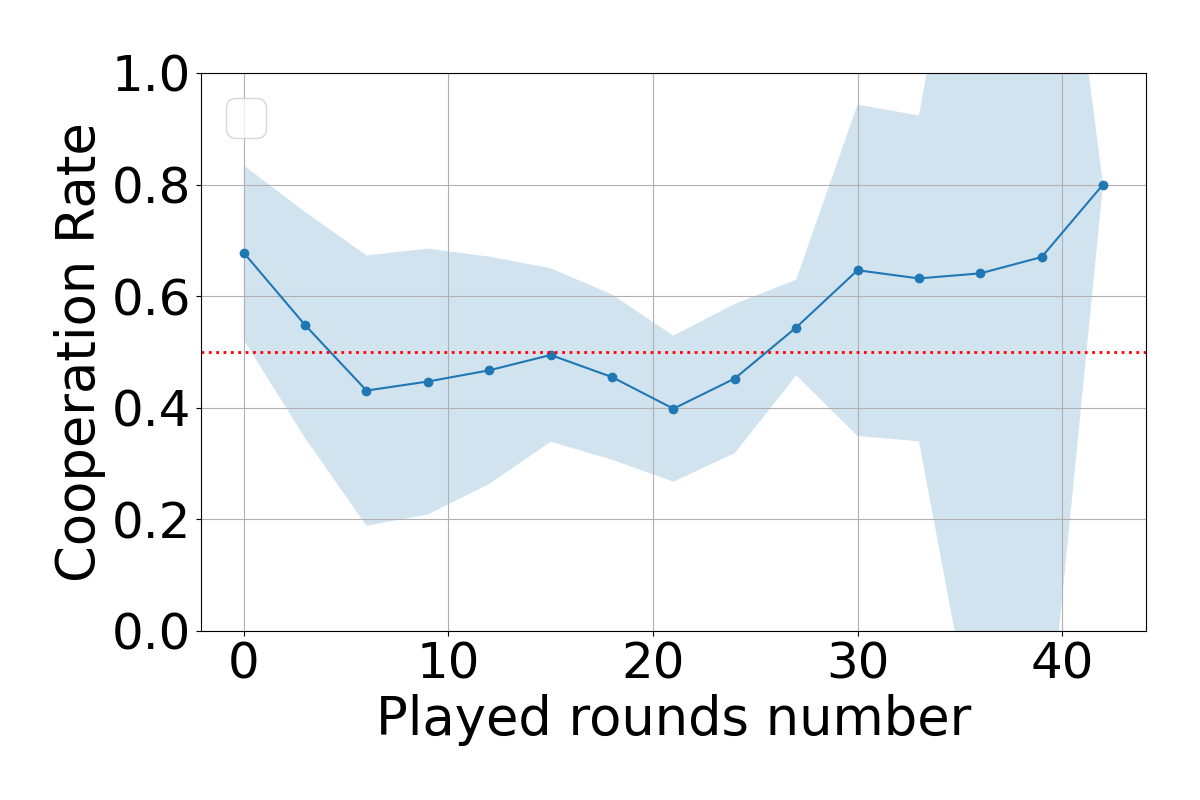}
        \caption{\(p_{c}\) Cogito, RI.}
        \label{fig:cogitoRI}
    \end{subfigure}
    \hfill
    \begin{subfigure}[b]{0.325\linewidth}
        \centering
        \includegraphics[width=\linewidth]{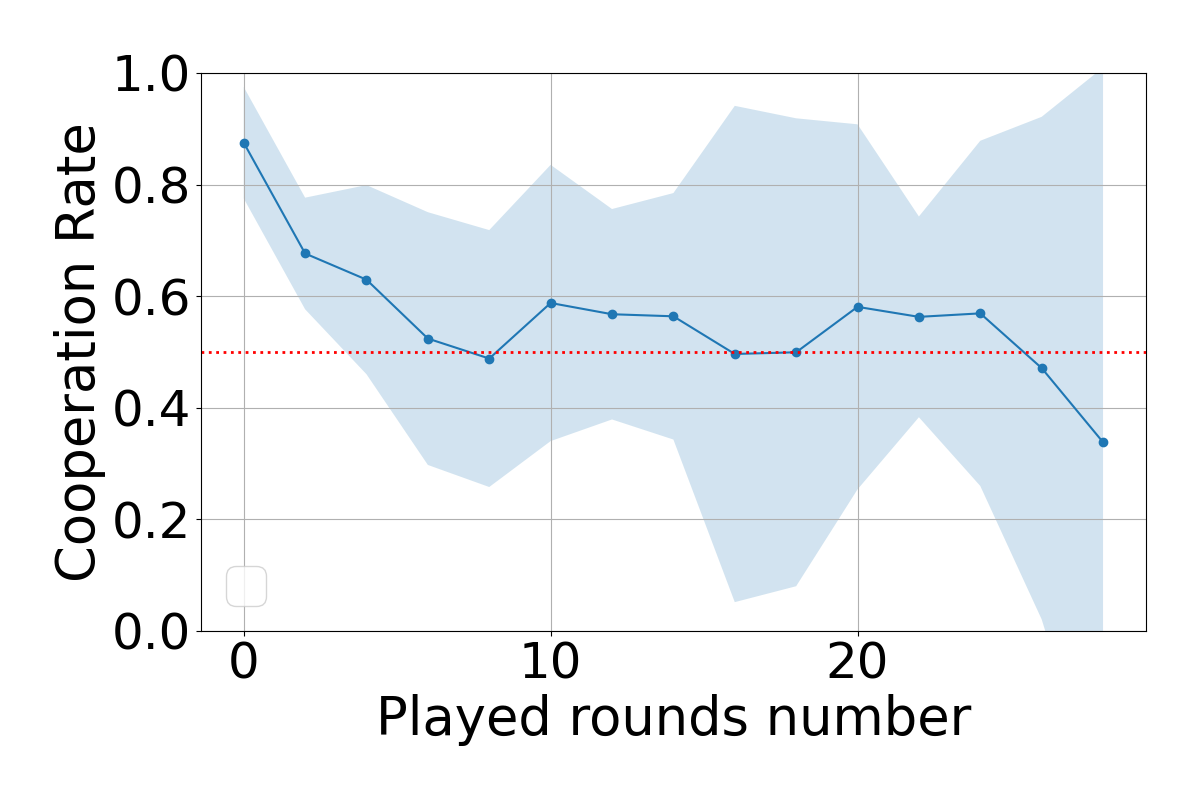}
        \caption{\(p_{c}\) Cogito, GC.}
        \label{fig:cogitoGC}
    \end{subfigure}
    \hfill
    \begin{subfigure}[b]{0.325\linewidth}
        \centering
        \includegraphics[width=\linewidth]{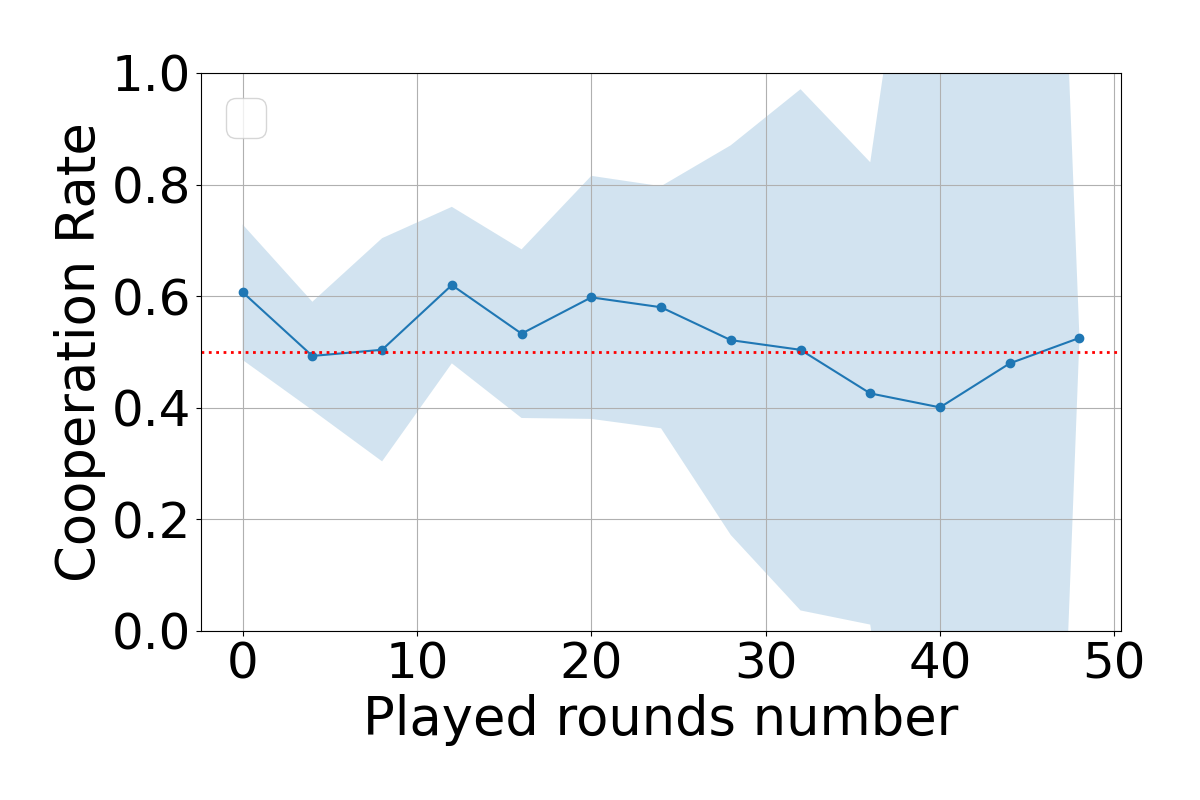}
        \caption{\(p_{c}\) Cogito, SA.}
        \label{fig:cogitoSA}
    \end{subfigure}

    \caption{Each plot shows the cooperation rate (\(p_{c}\)) evolution over the rounds played, for every LLM we show the results in the different tournament structures.}
    \label{fig:coop_rates}
\end{figure*}

The \texttt{Qwen 3} model (figures \ref{fig:qwen3RI}-\ref{fig:qwen3SA}) exhibits predominantly uncooperative behavior  in RI, maintaining a cooperation rate near zero in most rounds. When group dynamics are introduced, the model displays increased cooperative behavior, although this remains highly variable. \texttt{Phi 4} (figures \ref{fig:phi4RI}-\ref{fig:phi4SA}) on the other hand often exhibits a downward trend on cooperation rate, especially in RI. In the SA setting cooperation rates remain slightly higher.
Finally in \texttt{Cogito} (figures \ref{fig:cogitoRI}-\ref{fig:cogitoSA}) it is hard to draw clear trends, however cooperation is overall higher than the two previous models.

Tables~\ref{tab:qwen3_mu}-\ref{tab:cogito_mu} present the mean cooperation rates and corresponding 95\% confidence intervals for each model across different experimental conditions. \texttt{Cogito} consistently exhibits the highest mean values across all three conditions, with cooperation rates ranging from 0.50 (RI) to 0.55 (SA). This indicates a strong cooperative tendency, particularly in settings with group-based dynamics. In contrast, \texttt{Phi4} displays the lowest overall cooperation, especially in the Group Context. \texttt{Qwen3} demonstrates moderate behavior, with mean cooperation values clustered around 0.22–0.32 across conditions. In both \texttt{Phi4} and \texttt{Qwen3} the SA structure significantly boosts cooperation.

\begin{table*}
\centering

\begin{minipage}{0.32\textwidth}
\caption{\(\mu_{c}\) for \texttt{Qwen3}.}
\centering
\begin{tabular}{lcc}
\hline
\textbf{Qwen} & Mean & 95\% CI \\
\hline
RI & 0.22 & [0.20, 0.23] \\
GC & 0.23 & [0.20, 0.25] \\
SA & \textbf{0.32} & [0.29, 0.34] \\
\hline
\end{tabular}
\label{tab:qwen3_mu}
\end{minipage}
\hfill
\begin{minipage}{0.32\textwidth}
\caption{\(\mu_{c}\) for \texttt{Phi4}.}
\centering
\begin{tabular}{lcc}
\hline
\textbf{Phi} & Mean & 95\% CI \\
\hline
RI & 0.21 & [0.18, 0.24] \\
GC & 0.13 & [0.11, 0.16] \\
SA & \textbf{0.43} & [0.40, 0.46] \\
\hline
\end{tabular}
\label{tab:phi_mu}
\end{minipage}
\hfill
\begin{minipage}{0.32\textwidth}
\caption{\(\mu_{c}\) for \texttt{Cogito}.}

\centering
\begin{tabular}{lcc}
\hline
\textbf{Cogito} & Mean & 95\% CI \\
\hline
RI & 0.50 & [0.48, 0.53] \\
GC & \textbf{0.59} & [0.56, 0.62] \\
SA & 0.55 & [0.52, 0.57] \\
\hline
\end{tabular}
\label{tab:cogito_mu}
\end{minipage}

\end{table*}

\subsection{One-shot Cooperation}

\begin{figure*}
    \centering
    \begin{subfigure}[b]{0.325\linewidth}
        \centering
        \includegraphics[width=\linewidth]{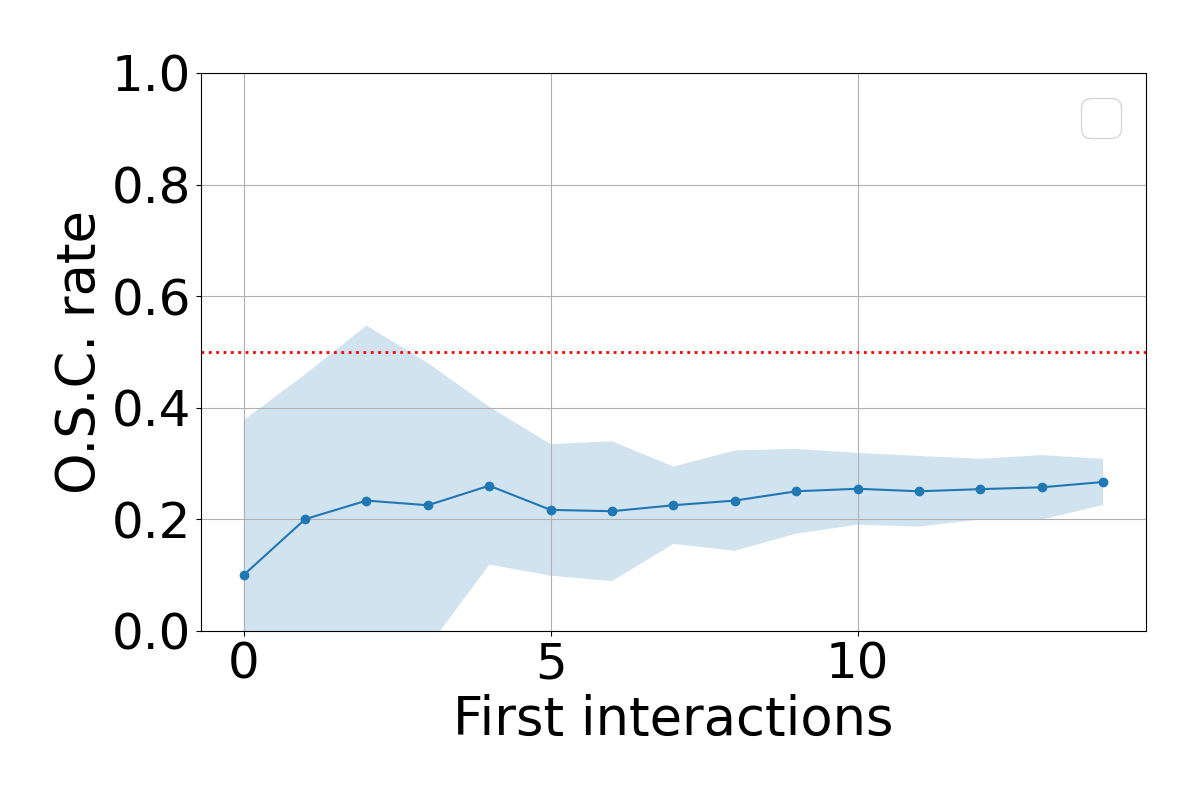}
        \caption{\(p_{osc}\) Qwen3, RI.}
        \label{fig:qwen3RI-osc}
    \end{subfigure}
    \hfill
    \begin{subfigure}[b]{0.325\linewidth}
        \centering
        \includegraphics[width=\linewidth]{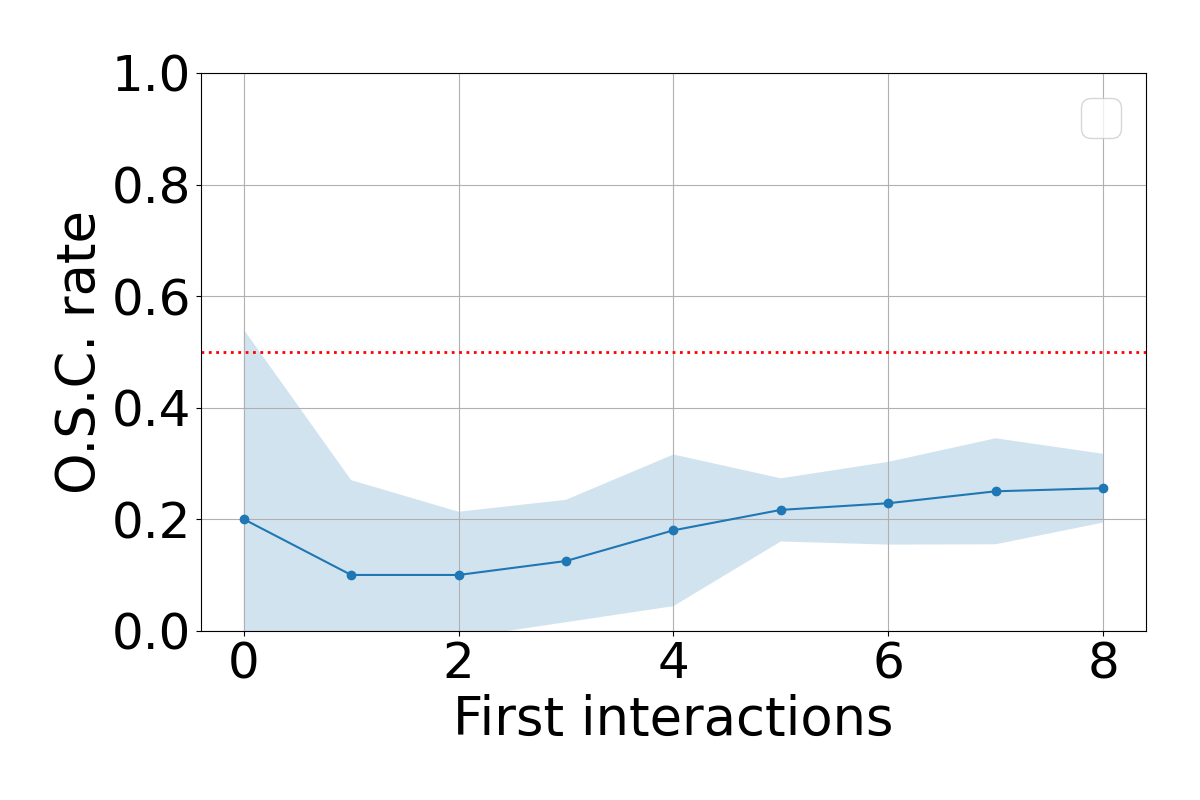}
        \caption{\(p_{osc}\) Qwen3, GC.}
        \label{fig:qwen3GC-osc}
    \end{subfigure}
    \hfill
    \begin{subfigure}[b]{0.325\linewidth}
        \centering
        \includegraphics[width=\linewidth]{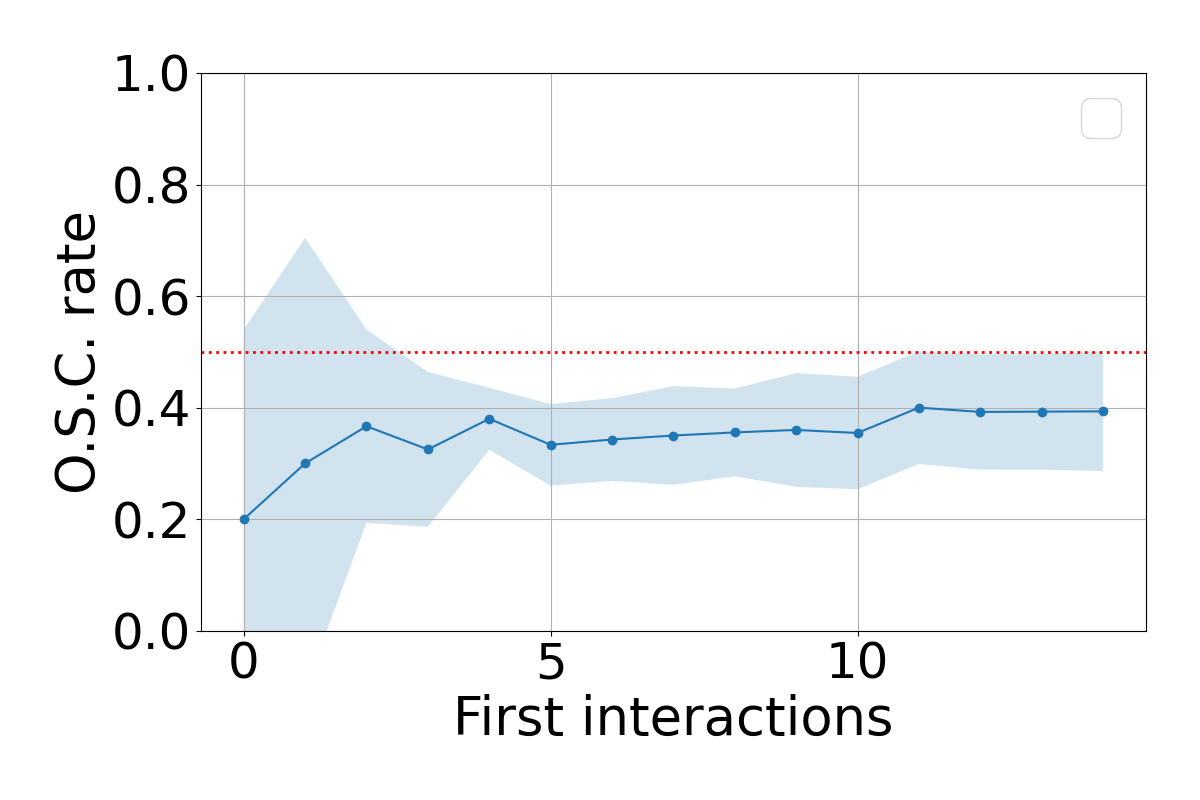}
        \caption{\(p_{osc}\) Qwen3, SA.}
        \label{fig:qwen3SA-osc}
    \end{subfigure}

    \vspace{0.5cm}

    \begin{subfigure}[b]{0.325\linewidth}
        \centering
        \includegraphics[width=\linewidth]{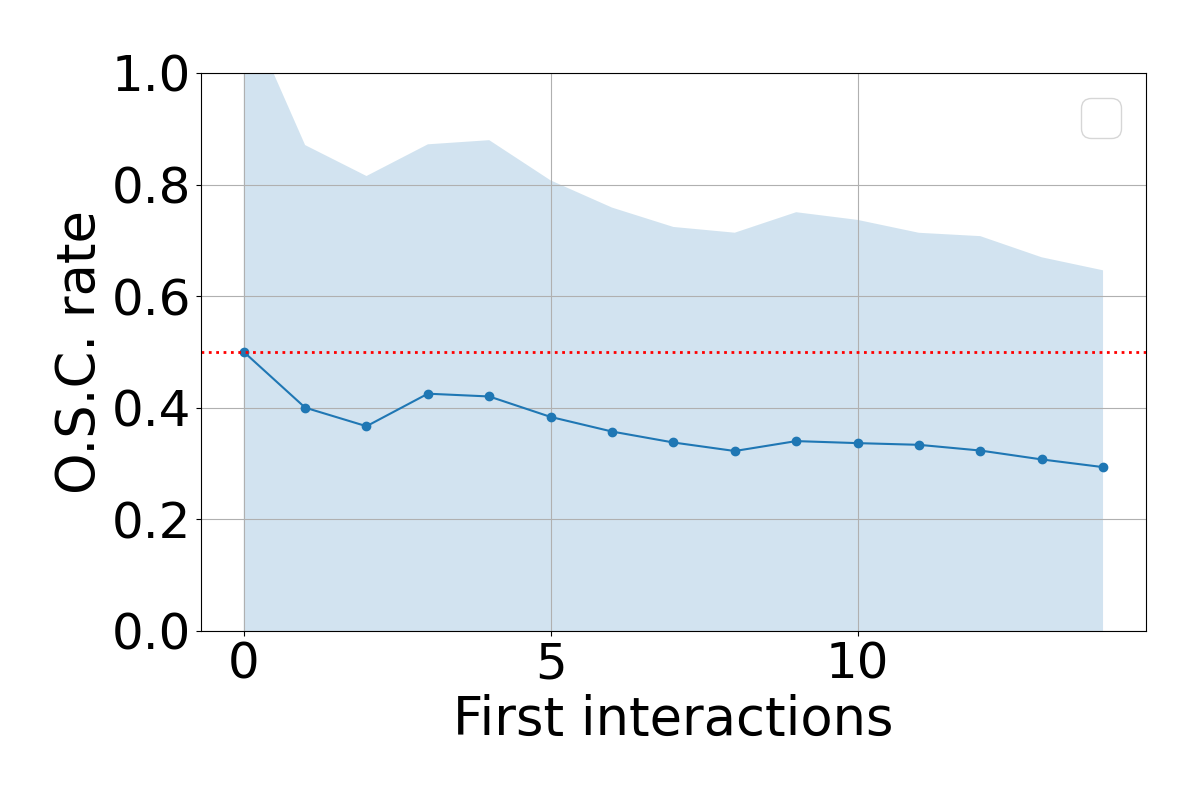}
        \caption{\(p_{osc}\) Phi4, RI.}
        \label{fig:phi4RI-osc}
    \end{subfigure}
    \hfill
    \begin{subfigure}[b]{0.325\linewidth}
        \centering
        \includegraphics[width=\linewidth]{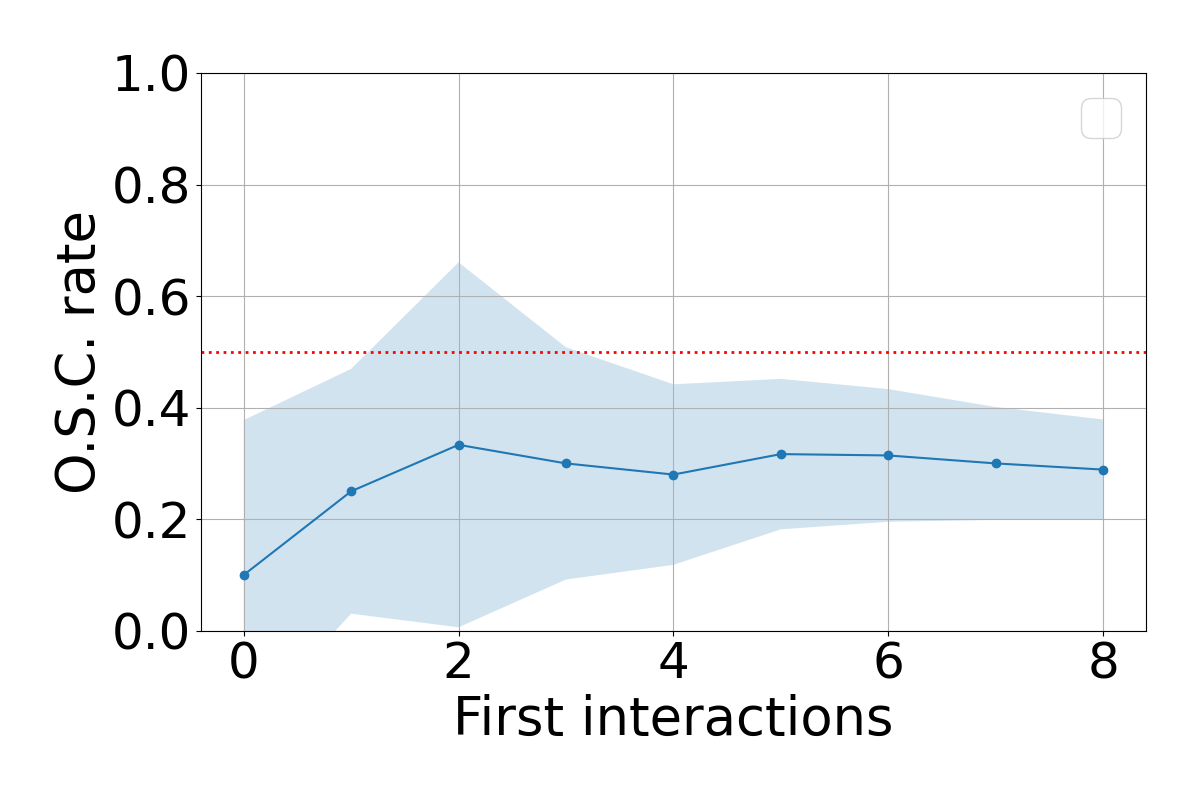}
        \caption{\(p_{osc}\) Phi4, GC.}
        \label{fig:phi4GC-osc}
    \end{subfigure}
    \hfill
    \begin{subfigure}[b]{0.325\linewidth}
        \centering
        \includegraphics[width=\linewidth]{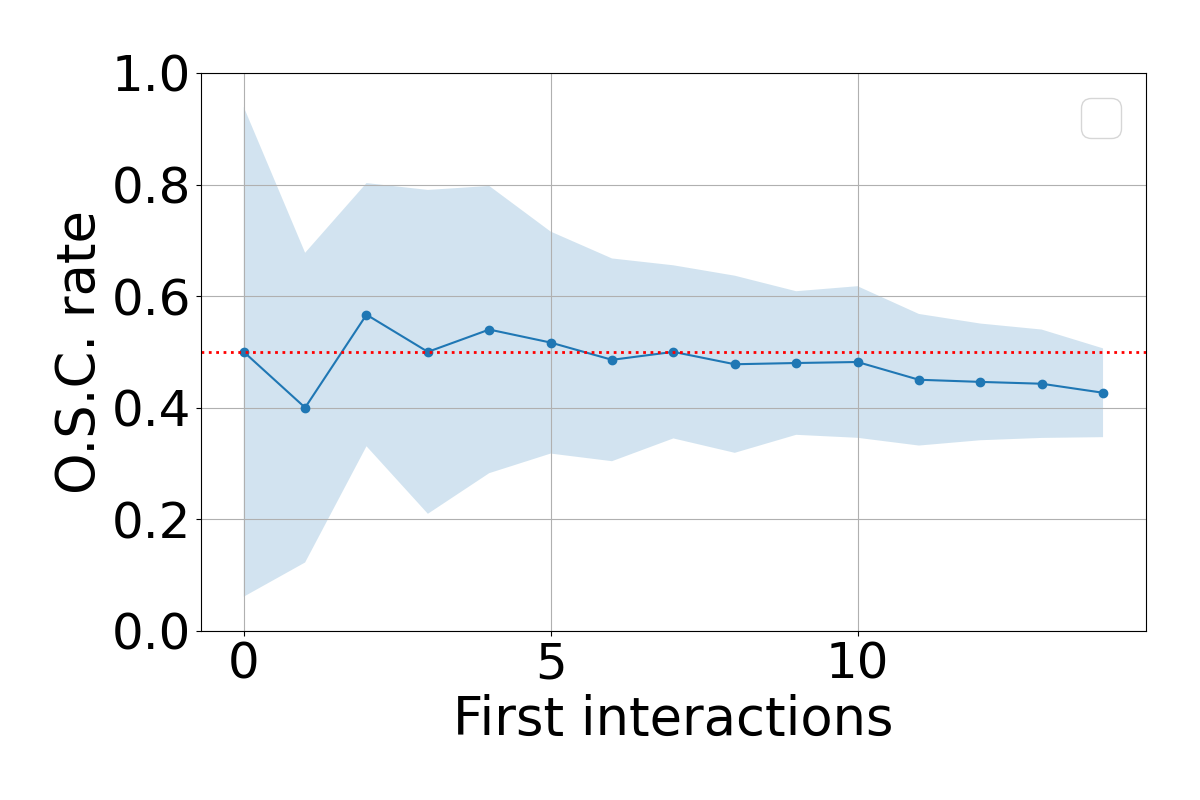}
        \caption{\(p_{osc}\) Phi4, SA.}
        \label{fig:phi4SA-osc}
    \end{subfigure}

    \vspace{0.5cm}

    \begin{subfigure}[b]{0.325\linewidth}
        \centering
        \includegraphics[width=\linewidth]{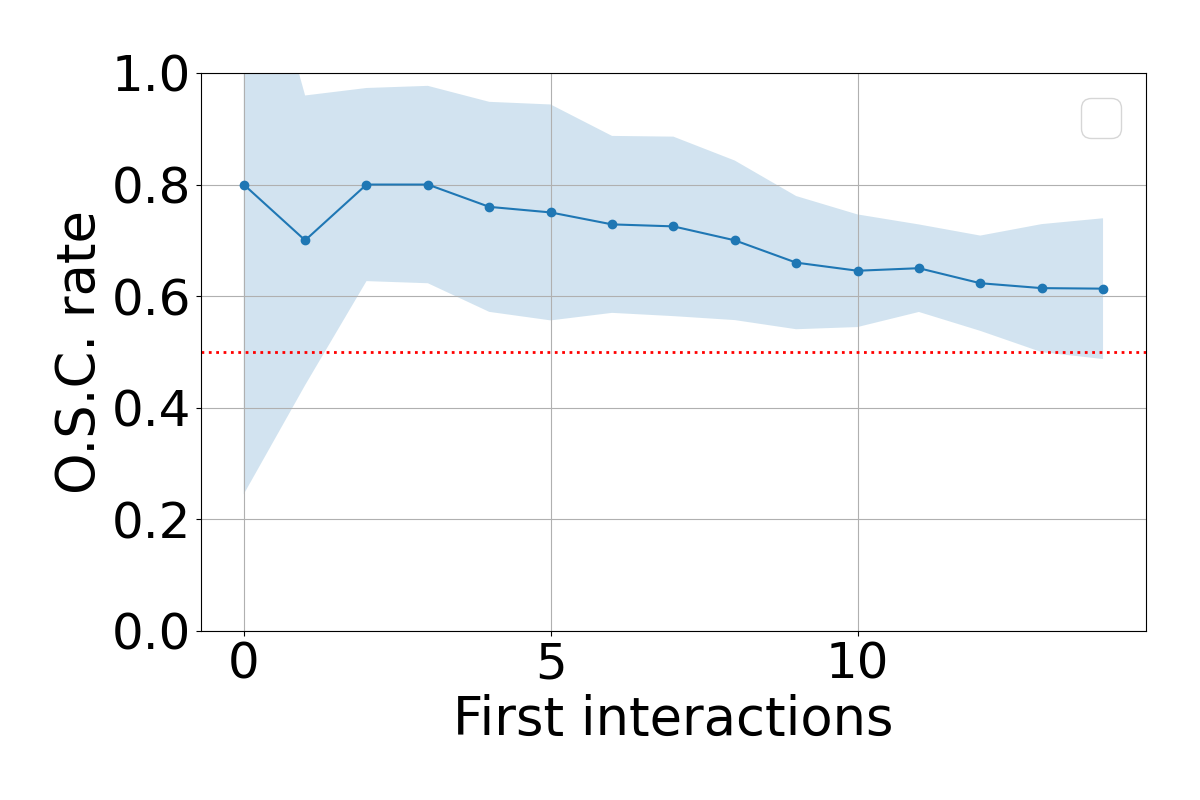}
        \caption{\(p_{osc}\) Cogito RI.}
        \label{fig:cogitoRI-osc}
    \end{subfigure}
    \hfill
    \begin{subfigure}[b]{0.325\linewidth}
        \centering
        \includegraphics[width=\linewidth]{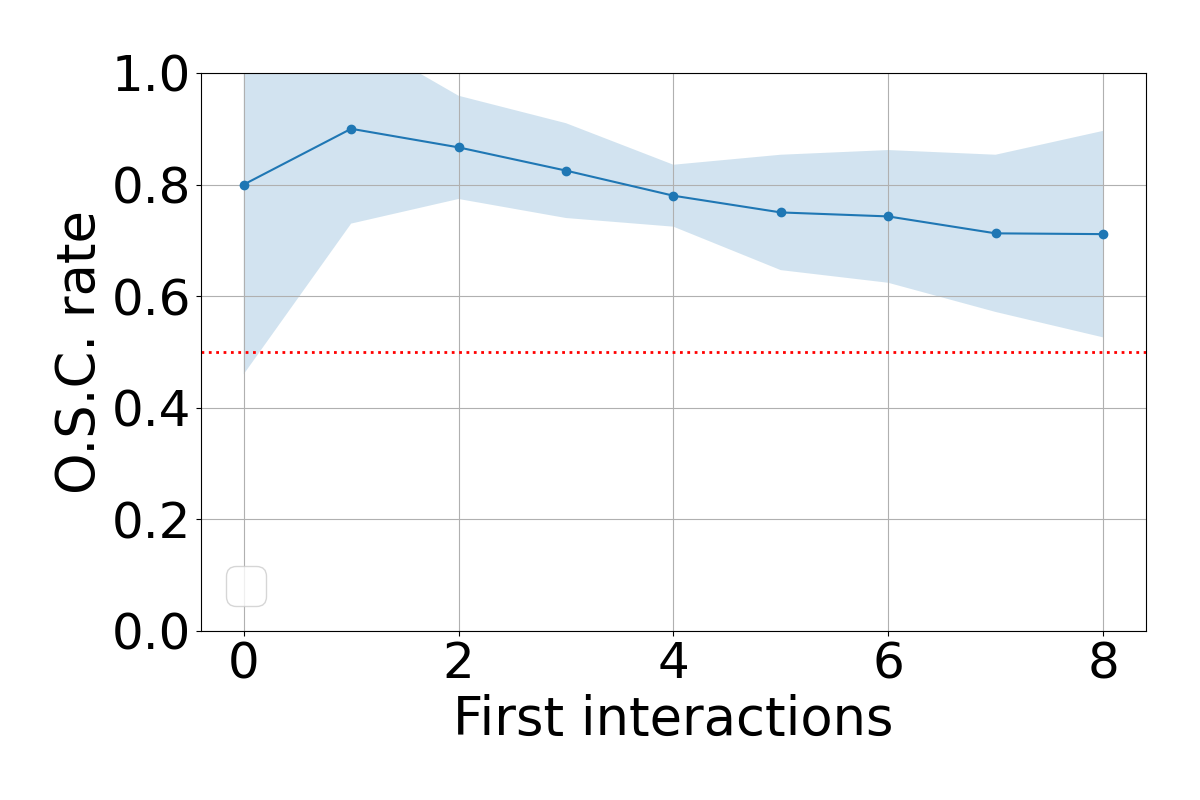}
        \caption{\(p_{osc}\) Cogito GC.}
        \label{fig:cogitoGC-osc}
    \end{subfigure}
    \hfill
    \begin{subfigure}[b]{0.325\linewidth}
        \centering
        \includegraphics[width=\linewidth]{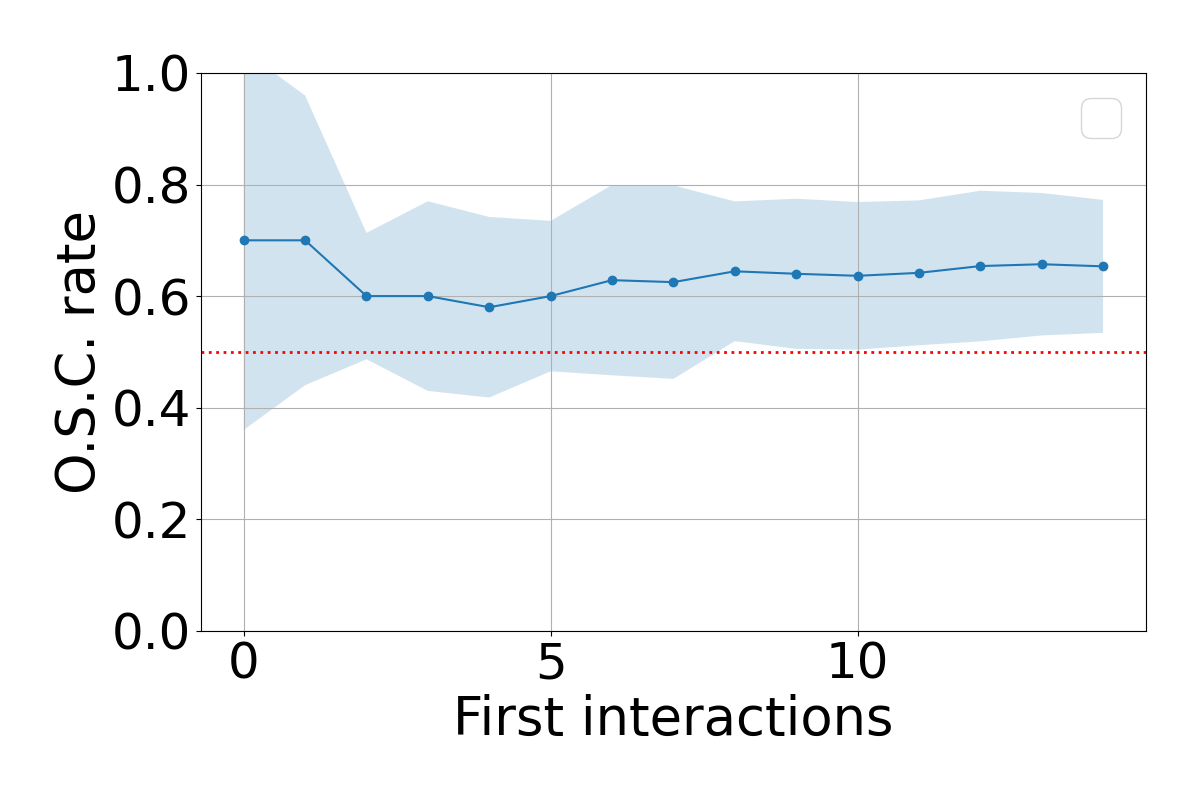}
        \caption{\(p_{osc}\) Cogito SA.}
        \label{fig:cogitoSA-osc}
    \end{subfigure}

    \caption{Each plot shows the one-shot cooperation rate (\(p_{osc}\)) evolution over the rounds played, for every LLM we show the results in the different tournament structures.}
    \label{fig:osc_rates}
\end{figure*}

In the one-shot interactions, \texttt{Qwen3} (figures \ref{fig:qwen3RI-osc}-\ref{fig:qwen3SA-osc}) exhibits relatively consistent behavior, particularly in the later rounds. Cooperation rates are higher in the super-additive setting. \texttt{Phi4} (figures \ref{fig:phi4RI-osc}-\ref{fig:phi4SA-osc}) follows a similar pattern with slightly higher one-shot cooperation with the SA structure.
\texttt{Cogito} (figures \ref{fig:cogitoRI-osc}-\ref{fig:cogitoSA-osc}) on the other hand seems to be more cooperative even on first interaction, however the highest cooperation is achieved in the group setting.

Tables~\ref{tab:qwen3_muosc}-\ref{tab:cogito_muosc} present the mean one-shot cooperation rates and corresponding 95\% confidence intervals for each model across different experimental conditions. \texttt{Cogito} remains the most cooperative, with notably high values in GC (0.71) and SA (0.65), suggesting a strong initial bias toward cooperation. \texttt{Phi4} and \texttt{Qwen3} again show lower values overall, with \texttt{Phi4} reaching its peak (0.43) in SA, similarly \texttt{Qwen3} attains its highest first-round cooperation (0.39) in SA.

\begin{table*}
\centering

\begin{minipage}{0.32\textwidth}
\caption{\(\mu_{osc}\) for \texttt{Qwen3}.}

\centering
\begin{tabular}{lcc}
\hline
\textbf{Qwen} & Mean & 95\% CI \\
\hline
RI & 0.27 & [0.20, 0.34] \\
GC & 0.26 & [0.16, 0.35] \\
SA & \textbf{0.39} & [0.31, 0.47] \\
\hline
\end{tabular}
\label{tab:qwen3_muosc}
\end{minipage}
\hfill
\begin{minipage}{0.32\textwidth}
\caption{\(\mu_{osc}\) for \texttt{Phi4}.}
\centering
\begin{tabular}{lcc}
\hline
\textbf{Phi} & Mean & 95\% CI \\
\hline
RI & 0.29 & [0.22, 0.37] \\
GC & 0.29 & [0.19, 0.38] \\
SA & \textbf{0.43} & [0.35, 0.51] \\
\hline
\end{tabular}
\label{tab:phi_muosc}
\end{minipage}
\hfill
\begin{minipage}{0.32\textwidth}
\caption{\(\mu_{osc}\) for \texttt{Cogito}.}
\centering
\begin{tabular}{lcc}
\hline
\textbf{Cogito} & Mean & 95\% CI \\
\hline
RI & 0.61 & [0.53, 0.69] \\
GC & \textbf{0.71} & [0.62, 0.81] \\
SA & 0.65 & [0.58, 0.73] \\
\hline
\end{tabular}
\label{tab:cogito_muosc}
\end{minipage}

\end{table*}

\subsection{Intra-group vs. Inter-group Cooperation}

\begin{figure*}
    \centering
    \begin{subfigure}[b]{0.325\linewidth}
        \centering
        \includegraphics[width=\linewidth]{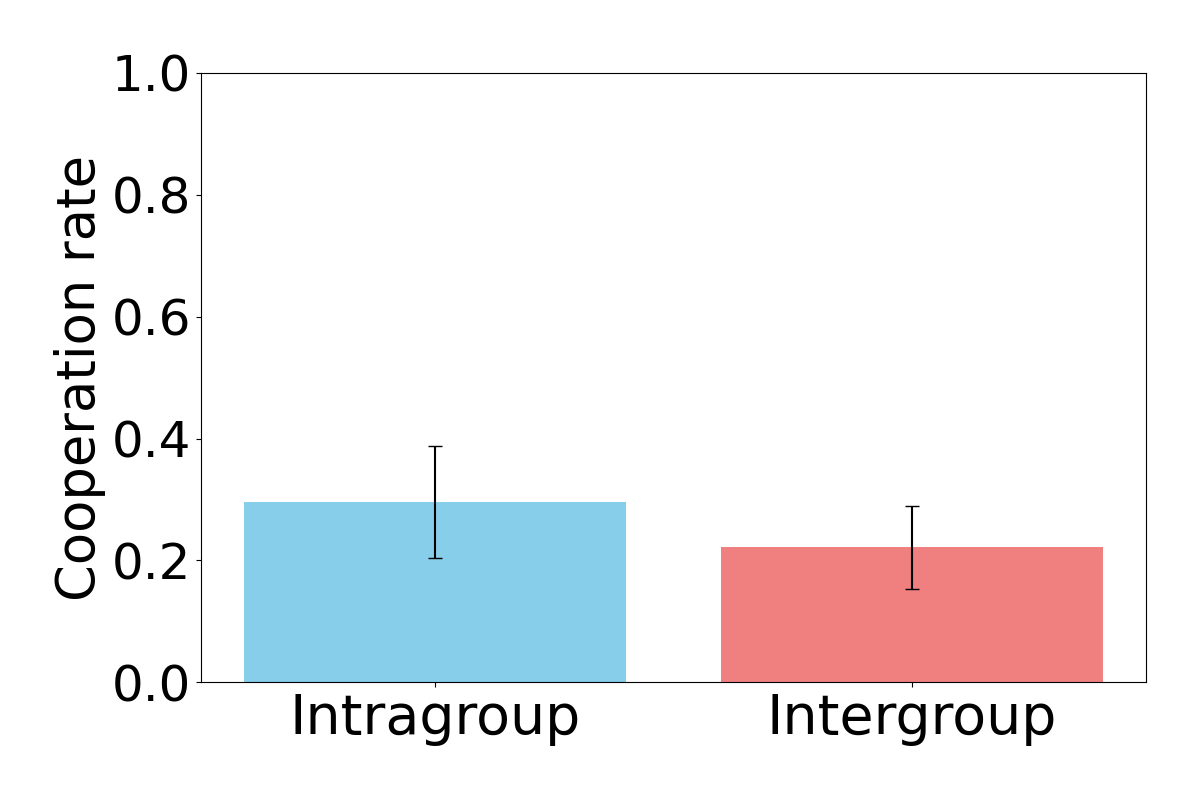}
        \caption{Qwen 3.}
        \label{fig:phi4RI}
    \end{subfigure}
    \hfill
    \begin{subfigure}[b]{0.325\linewidth}
        \centering
        \includegraphics[width=\linewidth]{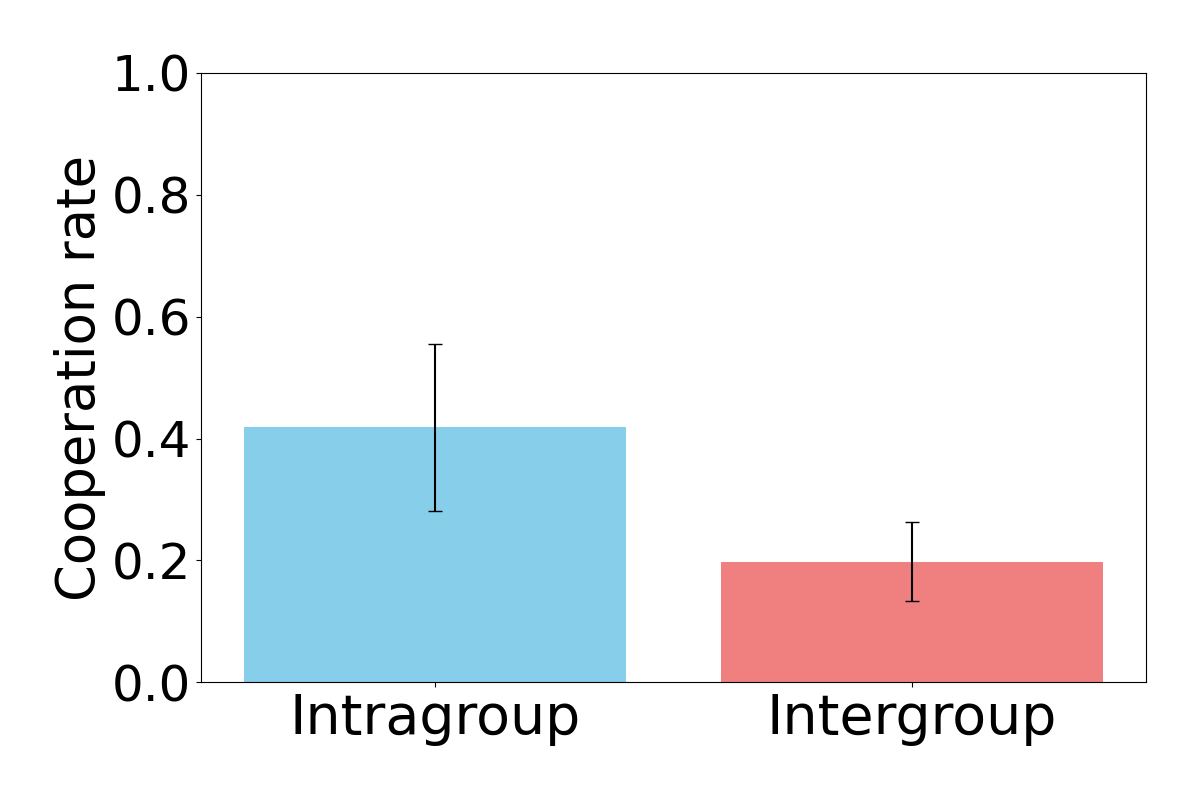}
        \caption{Phi 4.}
        \label{fig:phi4GC}
    \end{subfigure}
    \hfill
    \begin{subfigure}[b]{0.325\linewidth}
        \centering
        \includegraphics[width=\linewidth]{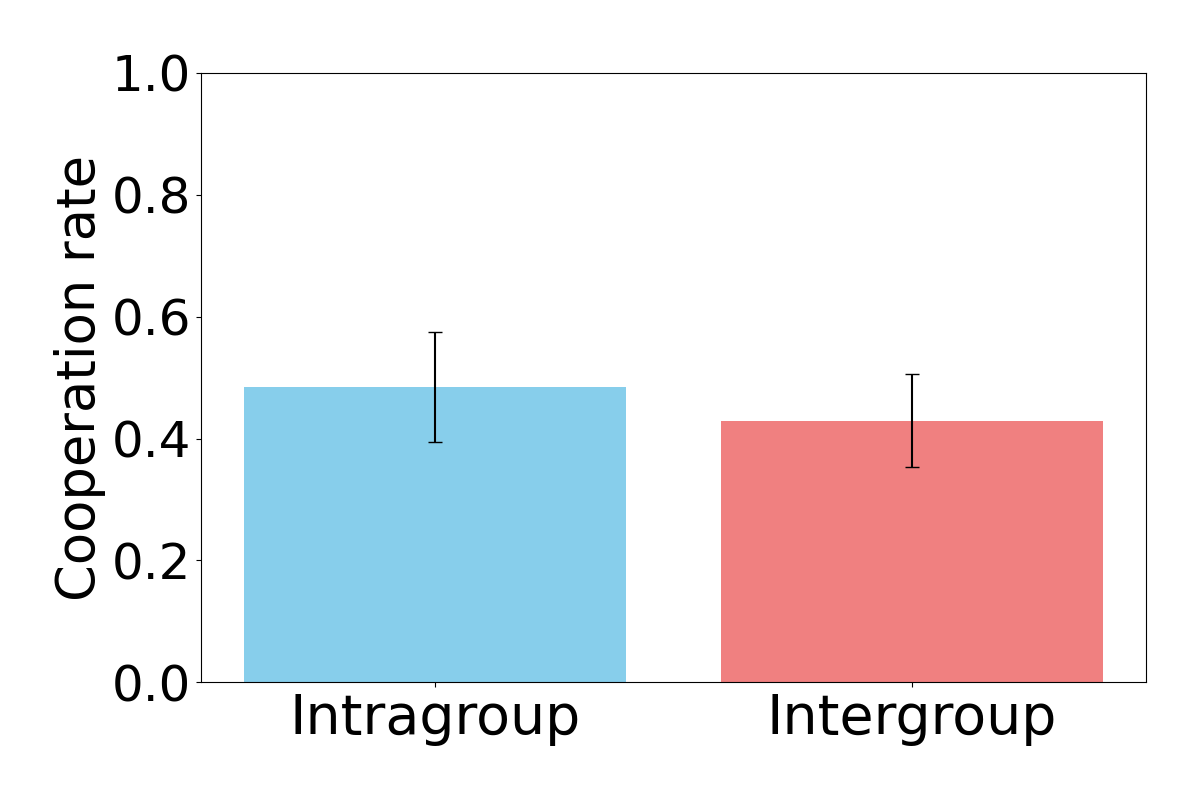}
        \caption{Cogito.}
        \label{fig:phi4SA}
    \end{subfigure}

    \caption{intra-group and inter-group cooperation in the super additive setting for the different models.}
    \label{fig:intervsintra}
\end{figure*}

Intra-group cooperation rates are higher (figure \ref{fig:intervsintra}), this is especially pronounced in \texttt{Phi4} with the highest difference. As noted before \texttt{Cogito} has a higher overall cooperation, however the difference between intra-group and inter-group cooperation is the lowest. 

\subsection{Results of Meta Prompting}

\begin{figure*}
    \centering
    \begin{subfigure}[b]{0.325\linewidth}
        \centering
        \includegraphics[width=\linewidth]{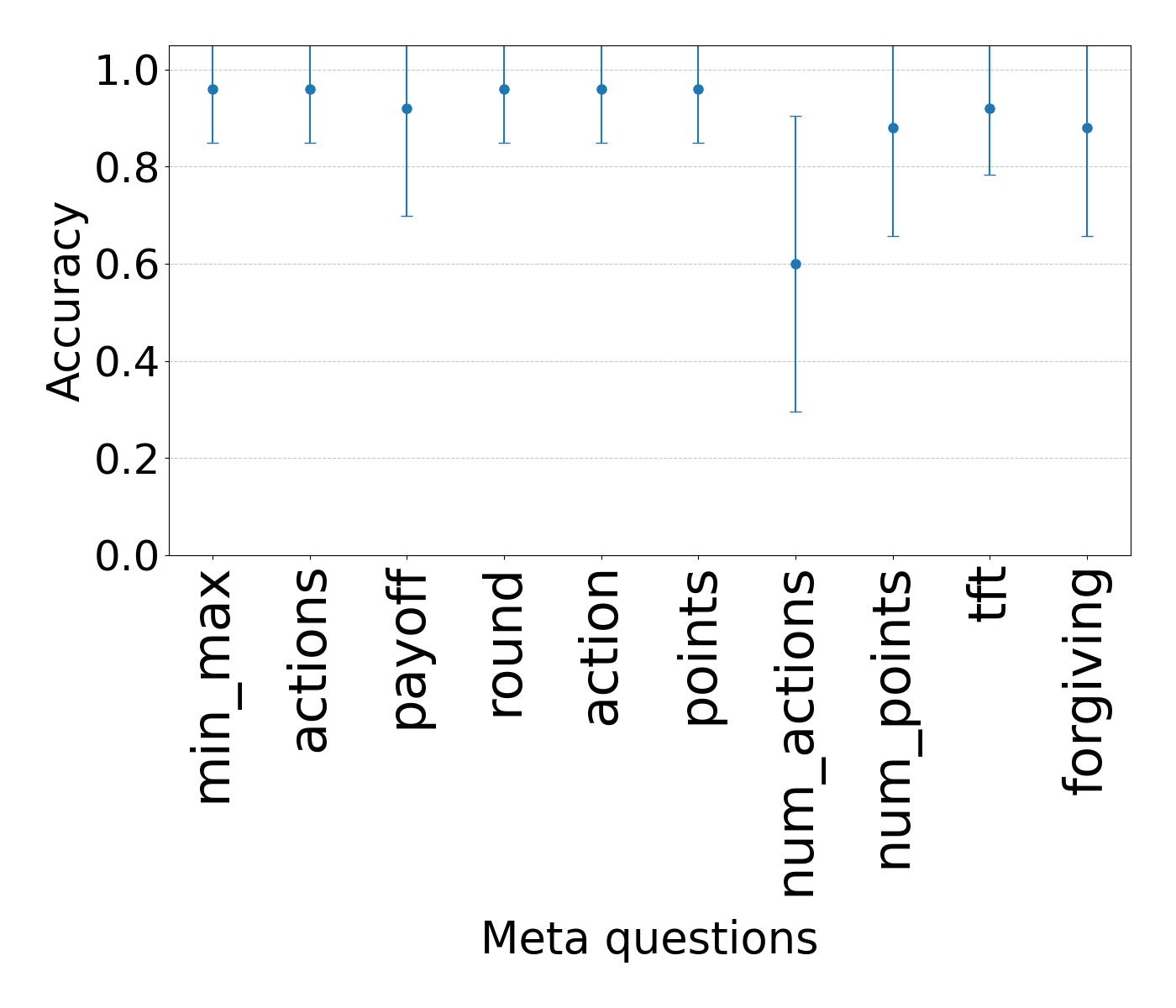}
        \caption{Qwen 3.}
        \label{fig:phi4RI}
    \end{subfigure}
    \hfill
    \begin{subfigure}[b]{0.325\linewidth}
        \centering
        \includegraphics[width=\linewidth]{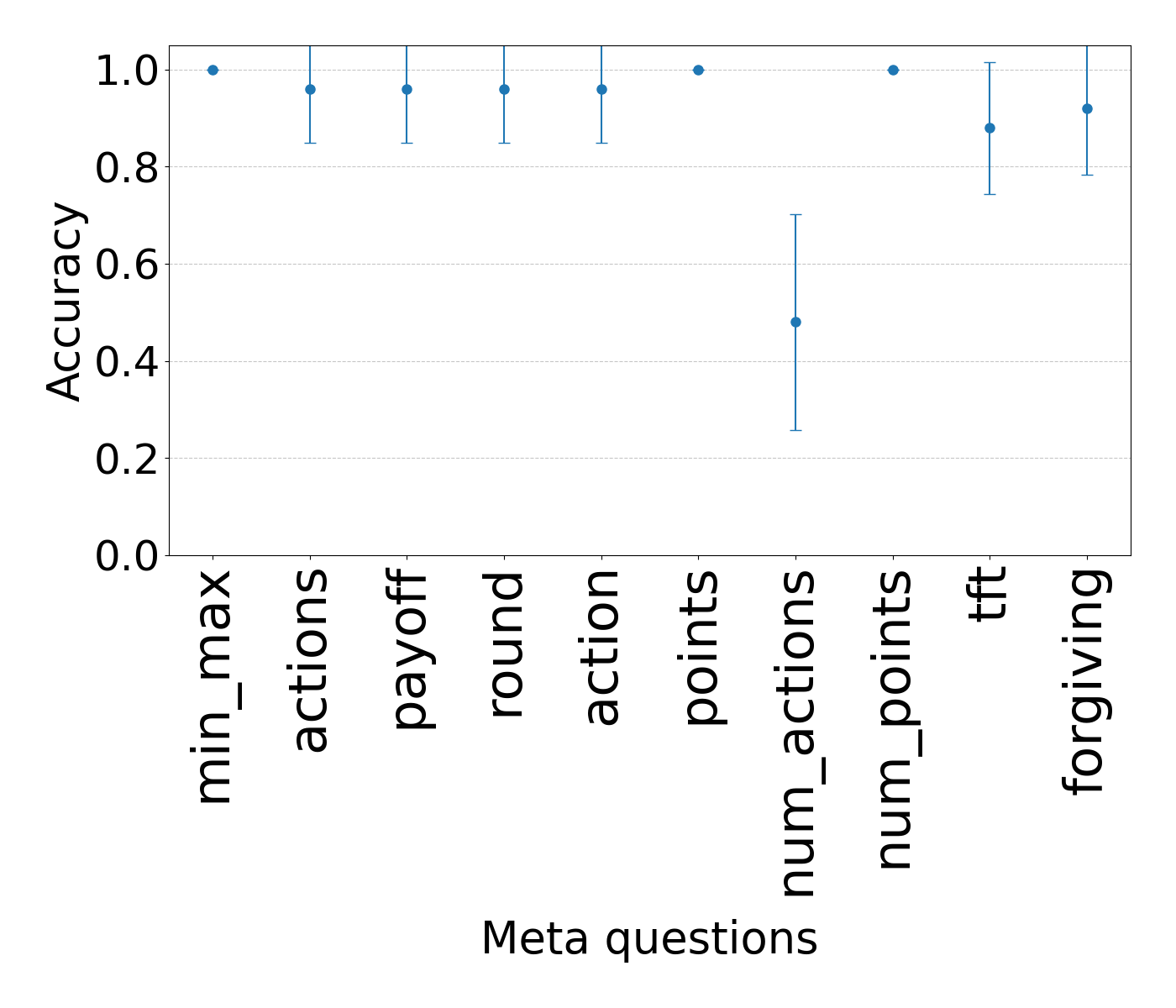}
        \caption{Phi 4.}
        \label{fig:phi4GC}
    \end{subfigure}
    \hfill
    \begin{subfigure}[b]{0.325\linewidth}
        \centering
        \includegraphics[width=\linewidth]{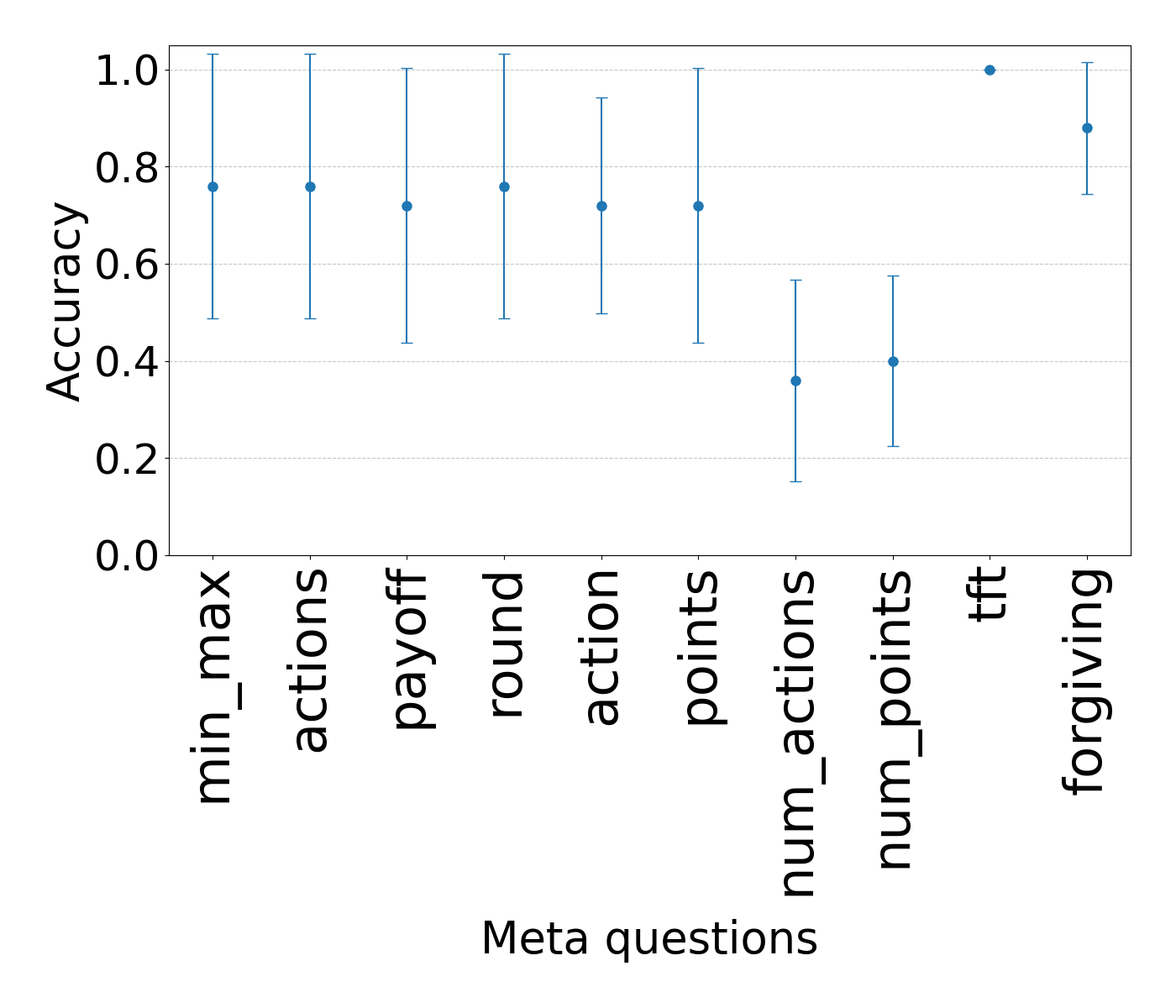}
        \caption{Cogito.}
        \label{fig:phi4SA}
    \end{subfigure}

    \caption{Meta prompt accuracy results from the SA tournaments for the different LLMs.}
    \label{fig:meta}
\end{figure*}

The meta-prompt questions serve as an indicator of how well each model understands the game rules, the current game state, and the behavior of the opponent (figure \ref{fig:meta}). Among the evaluated models, \texttt{Phi4} achieves the highest overall scores, followed closely by \texttt{Qwen3}. In contrast, \texttt{Cogito} exhibited lower performance, demonstrating a noticeable gap in performance relative to the other two models. These results suggest that \texttt{Cogito} lacks a sufficient understanding of the game dynamics, which in turn hinders its ability to play the game effectively.

\section{Discussion}
\label{sec:discussion}
This study investigated the emergence of super-additive cooperation among language model agents within a tournament-based iterated prisoner's dilemma framework. We examined how different conditions: repeated interactions, group competition, and their combination (super-additive); influenced cooperation rates, one-shot cooperation, and strategic behavior in three light-weight open-source LLMs: \texttt{Qwen3 14b}, \texttt{Phi4 reasoning}, and \texttt{Cogito 14b}.

Our results indicate varied responses across conditions. \texttt{Qwen3} generally exhibited low cooperation in RI and GC, with increased cooperation when group dynamics and individual interactions were combined. Its one-shot cooperation was also highest in the SA setting. \texttt{Phi4} often showed a declining trend in cooperation rates, particularly in RI, though SA conditions sustained higher cooperation. Similar to \texttt{Qwen3}, \texttt{Phi4}'s one-shot cooperation was most pronounced under the SA structure. \texttt{Cogito} displayed generally higher cooperation rates overall compared to \texttt{Qwen3} and \texttt{Phi4}, though clear trends were less discernible. One-shot cooperation peaked in the GC condition.

Regarding our hypotheses, the findings provide partial support. For $H_1$, which posited that the average cooperation rate \(\mu_c\) under the SA condition would exceed that under either RI or GC alone, the results for \texttt{Qwen3} (SA: 0.32 vs RI: 0.22, GC: 0.23) and \texttt{Phi4} (SA: 0.43 vs RI: 0.21, GC: 0.13) fully align with this hypothesis.
This is particularly pronounced in intra-group cooperation rates. A comparison of intra-group and inter-group cooperation rates reveals that the elevated overall cooperation observed in the SA condition is primarily due to intra-group interactions. In particular, \texttt{Phi4} and \texttt{Qwen3} exhibit notably higher cooperation rates in intra-group matches. This aligns with the findings of \cite{efferson2024superadditivecooperation}: introducing an external enemy through group competition increases pressure for intra-group cooperation.
In contrast, \texttt{Cogito} does not appear to consistently choose a more cooperative strategy in the SA setting. This is evidenced by the minimal difference between its intra-group and inter-group cooperation rates. Moreover, \texttt{Cogito} occasionally exhibits higher cooperation in the GC condition than in SA, further indicating a lack of strategic group play. This interpretation is supported by the meta-prompt evaluations, where \texttt{Phi4} and \texttt{Qwen3} demonstrated a better understanding of the game dynamics. \texttt{Cogito}’s weaker performance in these evaluations suggests that its behavior may be driven more by inherent model biases or superficial pattern matching than by deliberate strategic reasoning. The difference in behavior might also be caused by the model's training process, \texttt{Cogito} uses Iterated Distillation and Amplification \cite{christiano2018supervisingstronglearnersamplifying}.

$H_2$, which posited that average one-shot cooperation ($\mu_{osc}$) in the SA setting would exceed that of RI or GC alone, offers a particularly interesting insights. In one-shot interactions, agents lack prior knowledge of their opponent’s group. Despite this, elevated cooperation rates are still observed. \texttt{Qwen3} and \texttt{Phi4} exhibit the highest $\mu_{osc}$ in the SA condition, lending support to the hypothesis that super-additive structures can promote initial cooperative behavior. Interestingly, \texttt{Cogito} achieves its highest $\mu_{osc}$ in the GC condition, highlighting a divergence in model behavior.

Overall tournament scores and winners are not discussed for two primary reasons. First, because agents may exit a game at any point, they do not all participate in the same number of rounds, rendering direct score comparisons unfair. Second, the primary objective of this experiment is to evaluate collaboration dynamics rather than competitive performance. Consequently, we do not focus on fixed traditional IPD strategies or detailed game score analyses.

These results suggest that super-additive interaction structures can indeed enhance cooperation in large language models, the degree and consistency of this effect depend on the model's capacity to understand the environment dynamics. Nonetheless, for models with a strong understanding of game dynamics, combining RI with GC appears to promote more cooperative behavior during initial interactions. The SA condition thus emerges as a particularly effective framework for fostering cooperation and one-shot cooperation in language model agents.
Employing an interaction network structure that incorporates both repeated interactions and inter-group competition gives important insights on the counterintuitive role of competition in LLM alignment. Furthermore pressure from competing groups incentivizes intra-group cooperation, which in turn enhances the agents' overall success. Such a framework can enhance agent performance in domains characterized by a blend of cooperative and competitive dynamics, including financial negotiation, multiplayer online environments, and multi-robot systems in logistics. Pressure from competing groups incentivizes intra-group cooperation, which in turn enhances the agents' overall success.

It is noteworthy that some language models exhibit behavioral patterns similar to those observed in humans. In particular, several models showed especially high intra-group cooperation in the combined super-additive setting (\texttt{Phi4} and \texttt{Qwen3}), mirroring results from the Papua New Guinea experiment reported in \cite{efferson2024superadditivecooperation}. This highlights the potential of language models to simulate socially complex human behaviors.

\section{Limitations}
\label{sec:limitations}
While this study provides insights into LLM cooperation, several limitations should be acknowledged:
\begin{enumerate}
    \item \textbf{Scale:} The choice of smaller light-weight models was partly due to the computational demands of running numerous game simulations. Larger-scale experiments with more sophisticated models would require significantly more computational resources, which was a constraint for this study. The observed behaviors and the extent of super-additive cooperation may differ in more powerful proprietary models. Furthermore the tournament involved two teams of three players each; a larger number of agents, teams and experiment replications might reveal different emergent behaviors or more robust statistical patterns. The limited number of rounds per player might also constrain the full development of complex strategies. Finally, factors such as planning frequency, model temperature, and more prompting techniques should be systematically tested and compared.

    \item \textbf{Generalizability of Game Setting:} The study utilized the IPD as the primary interaction model. While IPD is a classic game for studying cooperation, the findings may not directly generalize to other types of social dilemmas, strategic games with different payoff structures, or more complex real-world interaction scenarios.

    \item \textbf{Prompt Sensitivity and Design:} LLM behavior can be highly sensitive to prompt phrasing. While efforts were made to create neutral and clear prompts, subtle interpretations by the models could still influence outcomes. The specific structure of the planner, evaluator, and player LLM prompts also shapes the observed behavior, this framework could not be extensively tested due to time constraints.

    \item \textbf{Fixed Model Parameters:} The study operated with fixed model parameters, without any fine-tuning during the experiments. This approach isolates context behavioral changes but does not explore how LLMs might adapt their underlying models or learn more deeply from experience over longer timescales.

    \item \textbf{Partner Choice Mechanism:} While players could terminate a match and proceed to a new opponent, the study did not deeply explore the nuances of more complex partner selection such as autonomous team selection, which might influence cooperation.
\end{enumerate}

\section{Conclusion and Future Work}
\label{sec:conclusion}
In conclusion, this study demonstrates that structural factors like repeated interactions and inter-group competition significantly influence LLM agent cooperation. The super-additive condition, combining both, often fosters higher cooperation and one-shot cooperation rates, particularly for models like \texttt{Qwen3} and \texttt{Phi4}. However, model-specific characteristics and comprehension levels play a crucial role in shaping these dynamics. Our framework provides a valuable tool for exploring complex social behaviors in LLM agents, contributing to the understanding of how autonomous AI systems might interact in multi-agent environments. The findings underscore the potential for designing interaction structures that promote cooperative outcomes, which is vital for aligning AI behavior with human values.
Future research could address these limitations by expanding the range of LLMs tested, exploring different game structures, increasing the scale of simulations, and investigating the impact of adaptive learning mechanisms and more sophisticated prompting strategies. Additionally, incorporating richer and more complex social dynamics, such as communication between agents, autonomous team choice or a reputation system could provide deeper insights into the emergence of cooperation. Furthermore, given the rapid evolution of language models, their cooperative behaviors must be continuously evaluated to track changes and emerging patterns.

\subsubsection*{Acknowledgments}
We have made use of generative AI systems such as \texttt{ChatGPT 4.5} and \texttt{Gemini 2.5 pro} for paraphrasing and increasing clarity of the report.

\bibliography{custom}
\bibliographystyle{splncs04}
\end{document}